%% file: arxiv.tex
\documentclass[10pt,twocolumn,letterpaper]{article}

\usepackage[pagenumbers]{cvpr} % To force page numbers, e.g. for an arXiv version
\usepackage{graphicx}
\usepackage{amsmath}
\usepackage{amssymb}
\usepackage{booktabs}
\usepackage{amsfonts}       % blackboard math symbols
\usepackage{nicefrac}       % compact symbols for 1/2, etc.
\usepackage{microtype}      % microtypography
\usepackage[dvipsnames]{xcolor}
\usepackage{multirow}
\usepackage{subcaption}
\usepackage{amssymb}% http://ctan.org/pkg/amssymb
\usepackage{pifont}% http://ctan.org/pkg/pifont
\usepackage{xr}
\usepackage[accsupp]{axessibility}
\usepackage[dvipsnames]{xcolor}
\usepackage{xr}
\usepackage{xr-hyper}
\usepackage{diagbox}

\newcommand{\cmark}{\ding{52}}%
\newcommand{\xmark}{\ding{56}}%
\newcommand{\fullcircle}{\ding{108}}%
\newcommand{\greencheck}{{\color{green}\cmark}}
\newcommand{\redx}{{\color{red}\xmark}}
\newcommand{\starrr}{\ding{72}}%
\newcommand{\emptycircle}{\ding{109}}%
\makeatletter

\usepackage[pagebackref,breaklinks,colorlinks]{hyperref}

\usepackage[capitalize]{cleveref}
\crefname{section}{Sec.}{Secs.}
\Crefname{section}{Section}{Sections}
\Crefname{table}{Table}{Tables}
\crefname{table}{Tab.}{Tabs.}

\def\cvprPaperID{6138} % *** Enter the CVPR Paper ID here
\def\confName{CVPR}
\def\confYear{2023}
\def\proposed{RNR-Map\xspace}
\def\proposedfullname{renderable neural radiance map\xspace}
\newcommand{\customspace}{\hspace*{15pt}}
\def\etc{\emph{etc}\onedot} 
\begin{document}

\title{Renderable Neural Radiance Map for Visual Navigation}

% \author{Obin Kwon \\
% Seoul National University \\ 
% {\tt\small obin.kwon@rllab.snu.ac.kr}
% \and
% Jeongho Park\\
% Seoul National University \\ 
% {\tt\small jeongho.park@rllab.snu.ac.kr}
% \and
% Songhwai Oh \\\
% Seoul National University \\ 
% {\tt\small songhwai@snu.ac.kr}

% }
\author{
Obin Kwon \customspace
Jeongho Park \customspace
Songhwai Oh 
\thanks{This work was supported by Institute of Information \& Communications Technology Planning \& Evaluation (IITP) grant funded by the Korea government (MSIT) (No. 2019-0-01190, [SW Star Lab] Robot Learning: Efficient, Safe, and Socially-Acceptable Machine Learning). \textit{(Corresponding author: Songhwai Oh)}}
\customspace
\\
Department of Electrical and Computer Engineering, ASRI, Seoul National University
\\
{\tt\small obin.kwon@rllab.snu.ac.kr, jeongho.park@rllab.snu.ac.kr, songhwai@snu.ac.kr}
\\}
\maketitle

%%%%%%%%% ABSTRACT
\begin{abstract}
We propose a novel type of map for visual navigation, a \proposedfullname (\proposed ), which is designed to contain the overall visual information of a 3D environment. 
The \proposed has a grid form and consists of latent codes at each pixel.
These latent codes are embedded from image observations, and can be converted to the neural radiance field which enables image rendering given a camera pose.
%
%The \proposed records the observed latent codes into corresponding grid cells.
%
The recorded latent codes implicitly contain visual information about the environment, which makes the \proposed visually descriptive.
This visual information in \proposed can be a useful guideline for visual localization and navigation. 
%
%Also, the renderable property of \proposed helps the agent to localize itself based on the observations.
%
We develop localization and navigation frameworks that can effectively utilize the \proposed.
We evaluate the proposed frameworks on camera tracking, visual localization, and image-goal navigation.
Experimental results show that the \proposed-based localization framework can find the target location based on a single query image with fast speed and competitive accuracy compared to other baselines.
Also, this localization framework is robust to environmental changes, and even finds the most visually similar places when a query image from a different environment is given.  
The proposed navigation framework outperforms the existing image-goal navigation methods in difficult scenarios, under odometry and actuation noises.
The navigation framework shows 65.7\% success rate in curved scenarios of the NRNS\cite{NRNS} dataset, which is an improvement of 18.6\% over the current state-of-the-art. 
Project page: \url{https://rllab-snu.github.io/projects/RNR-Map/}
\end{abstract}

\vspace{-0.7cm}
\section{Introduction}
In this paper, we address how to explicitly embed the visual information from a 3D environment into a grid form and how to use it for visual navigation. 
We present \textit{\proposedfullname(\proposed )}, a novel type of a grid map for navigation.%, which contains visual information about the 3D environment. 
We point out three main properties of \proposed which make \proposed navigation-friendly.
First, it is \textit{visually descriptive}.
Commonly used grid-based maps such as occupancy maps \cite{occ_ans, occ_lep, occ_nav} and semantic maps \cite{sem:active, sem:MultiON, semexp}, record obstacle information or object information into grids.
In contrast, \proposed converts image observations to latent codes which are then embedded in grid cells. 
Each latent code in a grid cell can be converted to a neural radiance field, which can render the corresponding region. 
We can utilize the implicit visual information of these latent codes to understand and reason about the observed environment.
For example, we can locate places based on an image or determine which region is the most related to a given image.
\proposed enables image-based localization only with a simple forward pass in a neural network, by directly utilizing the latent codes without rendering images.
We build a navigation framework with \proposed, to navigate to the most plausible place given a query image.
Through extensive experiments, we validate that the latent codes can serve as important visual clues for both image-based localization and image-goal navigation.
More importantly, a user has an option to utilize the renderable property of \proposed  for more fine-level of localization such as camera tracking.
% %

\proposed is \textit{generalizable}.
There have been a number of studies that leverage neural radiance fields (NeRF) for various applications other than novel view synthesis.
The robotic applications of NeRF are also now beginning to emerge \cite{imap, nice-slam, nerf-navi, nerf-rl, nerf-loc}. 
However, many of the approaches require pretrained neural radiance fields about a specific scene and are not generalizable to various scenes.
This can be a serious problem when it comes to visual navigation tasks, which typically assume that an agent performs the given task in an unseen environment \cite{embodied_survey}.
In contrast, \proposed is applicable in arbitrary scenes without additional optimization.
Even with the unseen environment, the \proposed can still embed the useful information from images to the map and render images.
The neural radiance fields of \proposed are conditioned on the latent codes.
A pair of encoder and decoder is trained to make these latent codes from images of arbitrary scenes and reconstruct images using neural radiance fields.
These pretrained encoder and decoder enable the generalization to unseen environments.

Third, \proposed  is \textit{real-time capable}. 
The majority of the present NeRF-based navigation methods require a significant time for inference because of required computation-heavy image rendering and rendering-based optimization steps.
The \proposed is designed to operate fast enough not to hinder the navigation system.
By directly utilizing the latent codes, we can eliminate the rendering step in mapping and localization. 
The mapping and image-based localization frameworks operate at 91.9Hz and 56.8Hz, respectively.
The only function which needs rendering-based optimization is camera tracking, which can localize under odometry noises, and it operates at 5Hz.

To the best of our knowledge, the \proposed is the first method having all three of the aforementioned characteristics as a navigation map.
The \proposed and its localization and navigation frameworks are evaluated in various visual navigation tasks, including camera tracking, image-based localization, and image-goal navigation.
Experimental results show that the proposed \proposed serves as an informative map for visual navigation. 
Our localization framework exhibits competitive localization accuracy and inference speed when compared to existing approaches.
On the image-goal navigation task, the navigation framework displays 65.7\% success rate in curved scenarios of the NRNS\cite{NRNS} dataset, where the current state-of-the-art method\cite{SLING} shows a success rate of 55.4\%.

\begin{figure*}[t]
\centering
\begin{subfigure}{0.73\textwidth}
\includegraphics[width=\textwidth, clip, trim=0.8cm 5.5cm 0.7cm 4.5cm ]{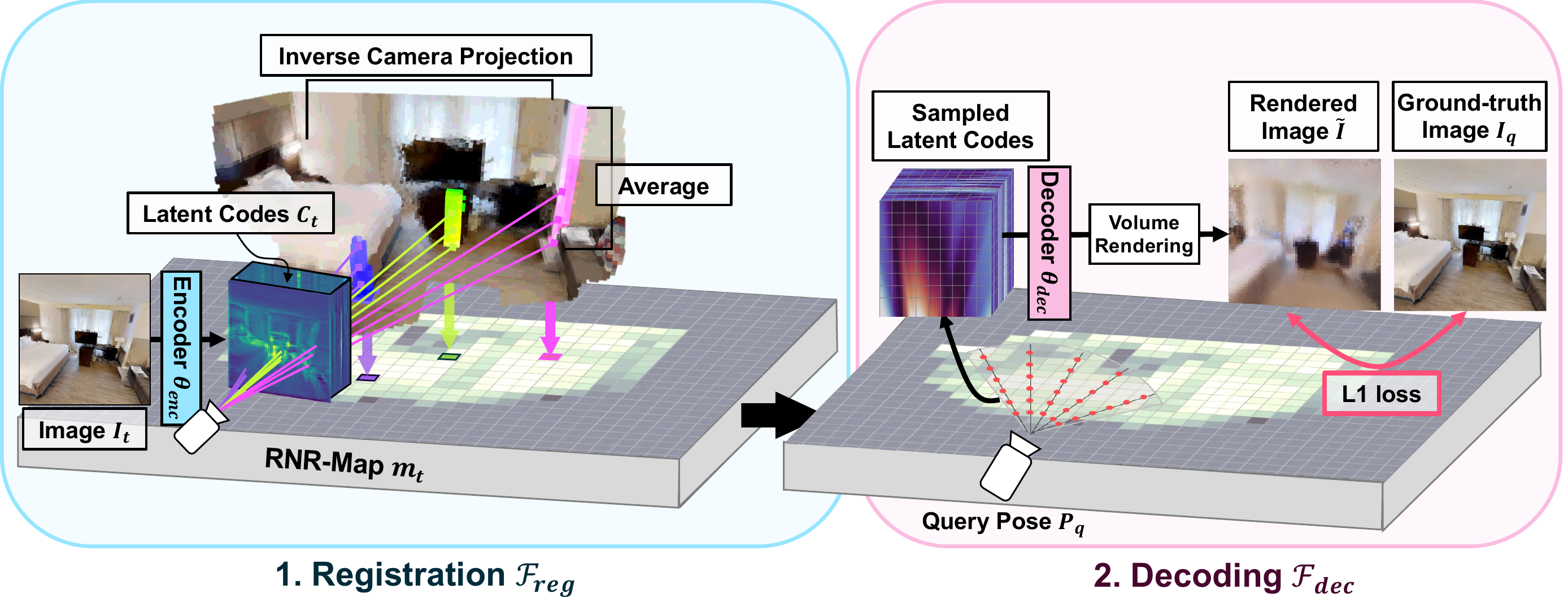}
\caption{\textbf{Reconstruction Framework}}
\label{fig:recon_overview}
\end{subfigure}%
\begin{subfigure}{0.269\textwidth}
  \includegraphics[width=1.1\textwidth, clip, trim=11.1cm 4.5cm 8.4cm 6.5cm ]{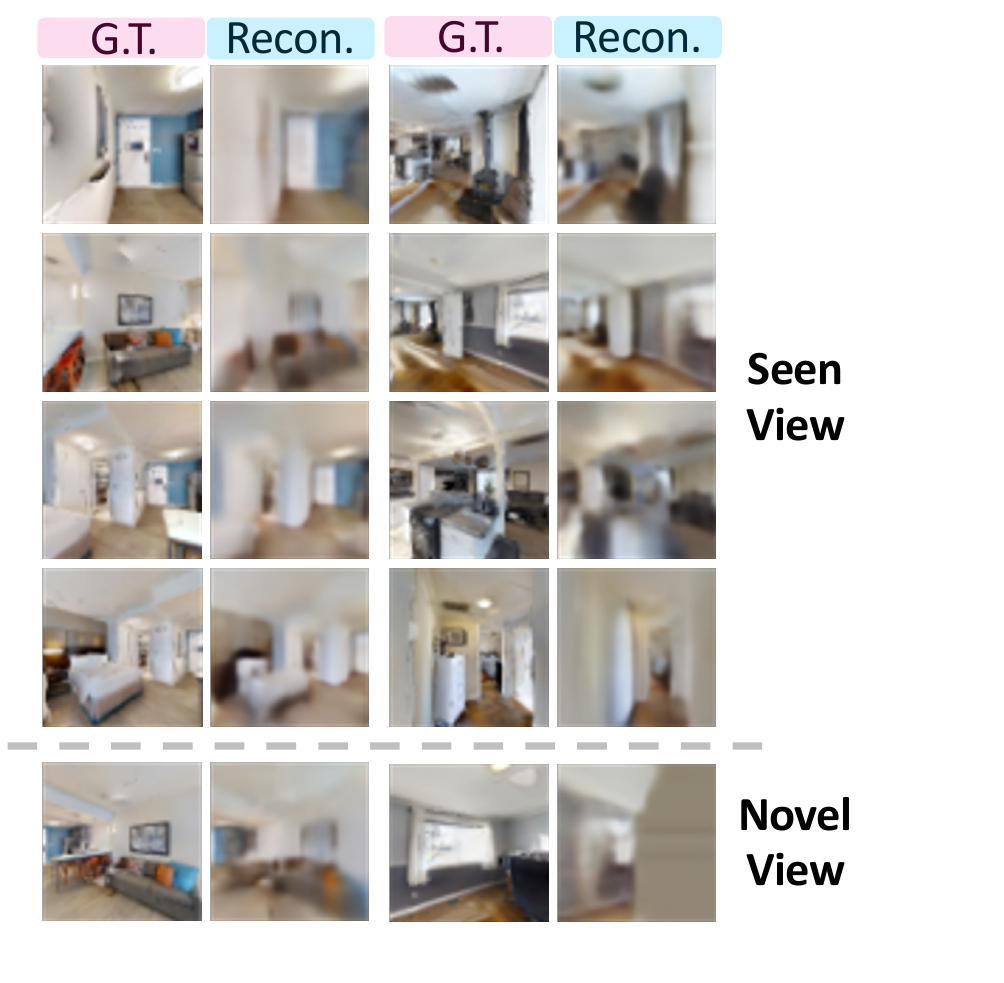}
  \caption{\textbf{Examples of the rendered images}}
\label{fig:recon_random}
\end{subfigure}
  \vspace{-0.5cm}
  \caption{(a) \textbf{Illustration of the reconstruction framework.} Two neural networks, encoder $\theta_{enc}$ and decoder $\theta_{dec}$ are used in this reconstruction framework.  (b) \textbf{Examples of the rendered images.} Odd columns are the given observations, and even columns are reconstructed results. The proposed method can reconstruct the images from the novel view (the last row).}
  \vspace{-0.5cm}
\end{figure*}

As \proposed finds a place based on the visual information of the map and the query image, we also consider a variant version of image-based localization.
In real-world scenarios, there can be partial changes in the target place (changes in furniture placement, lighting conditions, ...). 
Also, the user might only have images from similar but different environments.
We test the proposed localization framework in both cases.
We find that the \proposed is robust to environmental changes and is able to find the most visually similar places even when a novel query image from a different environment is provided.

The contributions of this paper can be summarized as follows:
\begin{itemize}
    \item We present \proposed, a novel type of renderable grid map for navigation, utilizing neural radiance fields for embedding the visual appearance of the environment.
    \item We demonstrate efficient and effective methods for utilizing the visual information in \proposed for searching an image goal by developing \proposed-based localization and navigation framework. 
    \item Extensive experiments show that the proposed method shows the state-of-the-art performance in both localization and image-goal navigation.
    %\item Novel formulation of image-goal navigation task, which is one step close to real-world scenarios.
\end{itemize}

\section{Related Work}
\paragraph{Embodied AI with spatial memories.}
One of the central issues in recent embodied AI research is how to construct a useful memory for the embodied agent  \cite{embodied_survey}.
A memory that contains the navigation history, as well as information about the observed environment, is required for successful task execution.
There is a large body of works using occupancy maps for visual navigation \cite{occ_nav, occ_ant, occ_ans, occ_lep}. 
An occupancy map expresses the environment in a grid form, and each grid cell has obstacle information about the corresponding region.
An occupancy map represents the overall structure of the environment and can guide a robot to navigate the environment safely.
There have been various kinds of research about building a spatial memory which contains additional information more than obstacles.  
The additional information can be object classes of the observed objects \cite{sem:active, sem:MultiON, semexp, sem:smnet}, or implicitly learned useful information for a specific task \cite{imp_CMP, imp_NeuralMap, MapNet, ISS}.
MapNet \cite{MapNet}, SMNet \cite{sem:smnet} and ISS \cite{ISS} have a similar mapping architecture with our method.
Like our approach, they learn how to convert RGBD observations into useful latent information, and record it in the spatial memories using 3D inverse camera projection. 
Using the recorded latent information, MapNet \cite{MapNet} learns to localize the agent pose, and SMNet learns to generate a semantic map. 
ISS \cite{ISS} is more related to ours since this method addresses scene generation and novel view image synthesis from a grid map.
Our research is focused on how we can utilize such latent information for visual navigation.
We develop localization and navigation frameworks which actively utilize the embedded visual information in \proposed.

\vspace{-0.5cm}

\paragraph{Robotics with neural radiance fields.}
The neural radiance field (NeRF) \cite{Nerf} has gained significant popularity in various AI tasks.
Not only in computer vision or graphics tasks, but NeRF is also adopted for robot applications in recent years.
NeRF predicts the RGB color and density of a point in a scene so that an image from an arbitrary viewpoint can be rendered.
This property enables pose estimation \cite{imap, barf, nerf-loc, nerf-navi} based on the photometric loss between the observed image and the rendered image, or manipulation of tricky objects \cite{nerf-dex, nerf-grasp2, nerf-dy, nerf-multi, nerf-grasp}.
A pretrained NeRF can also work as a virtual simulator, in which a robot can plan its trajectory \cite{nerf-navi} or can be used to train an action policy for the real-world \cite{nerf-sim2real}.
Among the existing approaches, NICE-SLAM \cite{nice-slam} is relevant to our work because it performs mapping and localization in arbitrary environments.
NICE-SLAM builds a 3D implicit representation of the environment from image observations. 
The camera pose is inferred from optimizing the photometric loss between the observation image and the rendered image.  
Our method, on the other hand, places less emphasis on mapping quality, and it is designed for successfully completing navigation tasks.
We focus on how the \proposed can be efficiently used for visual navigation, in terms of both speed and performance.
The mapping and the target-searching function of \proposed are designed to operate fast enough to not hinder the other parts of the system.
Also, the proposed \proposed method is generalizable in various environments without additional fine-tuning.

\section{\proposed } \label{sec:latent}
A robot agent has an RGB-D camera, and also knows its odometry information. 
Here, the odometry means how much the agent has moved from its previous position and we consider 3-DoF pose in this paper.
% 
%The robot agent takes the image observations along its trajectory.
%
At time $t$, the robot observes an RGBD image $I_t$ and its relative pose $\Delta p_t = (\Delta x_t,\Delta y_t,\Delta a_t)$ from the previous pose (xy position and heading angle).
By cumulating pose differences, the agent can determine its relative pose $p_{t}$ from the start pose $p_0=(0,0,0)$. 

A pair of the pretrained encoder and decoder is used when building a \proposed.
The training process of these encoder and decoder resembles the autoencoder method.
However, unlike 2D images, autoencoding a 3D environment is not a trivial problem. 
We build a reconstruction framework as shown in Figure \ref{fig:recon_overview}. 
%
% Two neural networks, the encoder and the decoder are used in the pretraining process.
%
The encoder encodes an image and embeds the pixel features into the \proposed.
We denote each embedded feature in a grid of \proposed as a latent code.
A query pose is then provided, and the decoder samples latent codes along each camera ray corresponding to each pixel and renders the corresponding images.
We present details of each part in the following section.
\vspace{-0.2cm}
\paragraph{Registration $\mathbf{F_\mathrm{reg}, F_\mathrm{map}}$.}
When an RGBD image $I_t \in \mathbb{R}^{H\times W\times 4}$ comes in, the encoder encodes the image into a same-sized feature $C_t \in \mathbb{R}^{H\times W\times D}$, where $H$ and $W$ refers to height and width of the image, respectively, and $D$ refers to the channel size.
First, each pixel $c_{h,w}\in \mathbb{R}^{D}$ in $C_t$ is mapped to its corresponding 3D world position $[q_x,q_y,q_z]^T$ using the known camera intrinsic $\bold{K}$, the extrinsic matrix $[\bold{R|t}]$ based on the agent pose, and the depth information $d_{h,w}$ of the pixel.
The world position of each pixel is calculated using inverse camera projection, as follows:
\begin{equation} \label{eq:inverse_projection}
\small
\begin{split}
\begin{bmatrix}
q_x(h,w) \\
q_y(h,w) \\
q_z(h,w) \\
\end{bmatrix} &= d_{h,w} \bold{R}^{-1} \bold{K}^{-1} 
\begin{bmatrix}
h \\
w \\
1 \\
\end{bmatrix} 
- \bold{t} \;.
\end{split}
\end{equation}
Then we digitize each position by normalizing with the map resolution $s$, and get map position $(u,v)$ as shown in (\ref{eq:map_pixel}). 
%(u, v) = \Bigl\lfloor{{x}\over{s}}  \Bigl\rceil, \Bigl\lfloor{{y}\over{s}} \Bigl\rceil$
%
We aggregate the pixel features that belong to the same 2D map position, and average the aggregated features into a single feature vector.
The pixel features are registered in the corresponding grid in the \proposed  $m\in\mathbb{R}^{U \times V \times D}$, where $U$ and $V$ refer to the height and width of the \proposed, respectively.
The number of averaged features at each 2D map position is also recorded in mask $n\in\mathbb{R}^{U\times V}$.
We denote the element of $m$ at the map position $(u,v)$ by $m(u,v)$ and the mask at $(u,v)$ by $n(u,v)$.
The registered feature $m(u,v)\in\mathbb{R}^{D}$ is the latent code which contains visual information of the region corresponding to the map position $(u,v)$.
This process can be written as follows:
\begin{align}\label{eq:map_pixel}
\small
    X_{(u,v)} &=  \Bigl\{ c_{h,w} \in C_t \big{|} u = \Bigl\lfloor{{q_x (h,w)}\over{s}}  \Bigr\rceil, v = \Bigl\lfloor{{q_y (h,w)}\over{s}} \Bigr\rceil  \Bigr\}  \notag \\
    m(u,v) &= {{1}\over{n(u,v)}} {\sum_{c_i \in X_{(u,v)}} \!\!\!c_i}, \quad n(u,v) = |X_{(u,v)}|. 
\end{align}
% %
% Note that these latent codes to 3D position mapping cannot be one-to-one mapping if multiple observations exist.
% %
% Moreover, we consider the \proposed  to have a 2.5D shape (height, width and feature dimension).
% %
% Therefore we average the latent codes in the same 2D position (averaging along with the height). 
%
The registration process $F_\mathrm{reg}$ includes the encoding of $C_t$, inverse projection, and feature registration.
The $F_\mathrm{reg}$ is represented as:%
\vspace{-0.3cm}
\begin{equation} \label{eq:F_reg}
\begin{split}
    m_t^{l}, n_t^{l} &= F_\mathrm{reg}(I_t, p_t; \theta_\mathrm{enc}),
\end{split}
\end{equation}
where $\theta_{enc}$ refers to the network parameters of the encoder.
The registration process $F_\mathrm{reg}$ outputs a local map $m_t^{l}$ and a local mask $n_t^{l}$.
The local map $m_t^{l}$ only contains the information from the image $I_t$, and this will be integrated with other local maps to form the global map $m^g$.\footnote{For simplicity, $m$ without any superscript refers to the global map ($m^g$) in the rest of the paper.}
%
%The superscript $l$ denotes that the outputs of $F_\mathrm{reg}$ are local, and only contain the information from the image $I_t$.
%
%This local map $w_t^{l}$ and mask $m_t^{l}$ will be integrated with other local maps and masks to form the global map $w^g$ and mask $m^g$.
%
%This mask $m$ is used when the next registration of the new image observation.
%
When multiple observations are given, we can use $n$ to compute the average value from the original and new latent codes.
We name this integration process $F_\mathrm{map}$, which operates $F_\mathrm{reg}$ over multiple observations.
$F_\mathrm{map}$ at time $t$ with previous $m^g_{t-1}$ is formulated as follows:
\begin{equation} \label{eq:F_map}
\begin{split}
    (m^g_t, n^g_t) &= F_\mathrm{map}(I_t, p_t, m^g_{t-1}, n^g_{t-1}; \theta_\mathrm{enc}) \\
    m^g_{t}(u,v) &= {{m^l_{t}(u,v) \cdot n^l_{t}(u,v) + m^g_{t-1}(u,v) \cdot n^g_{t-1}(u,v)}\over{n^l_t(u,v) + n^g_{t-1}(u,v)}} \\
    n^g_{t}(u,v) &= n^l_t(u,v) + n^g_{t-1}(u,v) \;.
\end{split}
\end{equation}

% %
% This latent code registration method is inspired by GRF \cite{GRF}. 
% %
% Although GRF uses an attention mechanism for better feature aggregation and their latent representation is 3D, we averaged along one dimension and made it 2D for simplicity.
% %
% We found that averaging is enough to produce the desired performance. 

\vspace{-0.8cm}
\paragraph{Decoding $\mathbf{F_\mathrm{dec}}$.}
To make these latent codes contain visual information, we reconstruct the image observation from the latent codes.
We use a decoder which has a structure similar to the generative scene network (GSN) \cite{GSN} for rendering an RGBD image from the 2D latent map.
Originally, GSN generates a random indoor floorplan from random variables. 
Then GSN proposed how to render images from the generated floorplan, based on locally conditioned radiance fields. 
Our approach involves designing the encoder $\theta_\mathrm{enc}$ and the registration process $F_\mathrm{reg}$ to transform image observations into latent codes, which can be converted to the locally conditioned radiance fields. 
We utilize the locally conditioned radiance fields from GSN to render an image from $m$.
Given the camera parameters, we can sample latent codes on points along the camera ray, corresponding to the pixel location which will be rendered in the camera.
%utilize e
The sampled latent codes are converted into modulation linear layer-based locally conditioned radiance fields \cite{CIPS, GSN}.
The decoder is trained to render an RGBD image from the latent codes to be close to the image observations $I_t$. 
The reader can refer to \cite{GSN} for a more detailed explanation about the rendering procedure. 

By training the rendered RGBD images $\tilde{I}_t$ to be close to the original observation $I_t$, the encoder is trained to extract the visual information from the image. 
This mechanism is similar to training an autoencoder, to extract and summarize useful information from an image into a latent vector.
We sample $N$ images in a random scene and embed them into the \proposed .
Then, we render each image from the final \proposed  and compare them with the original images.
The encoder and the decoder are trained using the following loss:
\begin{align} \label{eq:recon}
\small
    m^g_i, n^g_i &= F_\mathrm{map}(I_i, p_i, m^g_{i-1}, n^g_{i-1}; \theta_\mathrm{enc}), {\scriptstyle{i=1:N}} \notag \\
    \mathrm{Loss}(\theta_\mathrm{enc}, \theta_\mathrm{dec}) &= {{1}\over{N}} \sum_{i=1}^N ||I_i-F_\mathrm{dec}(m_N^g,p_i; \theta_\mathrm{dec})||_1,  
\end{align}
where $\theta_\mathrm{enc}$ and $\theta_\mathrm{dec}$ are weight parameters of the encoder and decoder, respectively.

% We want the \proposed  to express the overall 3D geometric structures of the environment, not only limited to the given observations.
% %
% This is one of the reasons why the \proposed  is trained using reconstruction instead of localization itself for training latent codes. 
% %
Since the rendering process is conditioned on the latent codes from the image observation, our proposed reconstruction framework is capable of embedding and rendering arbitrary images from unseen environments. This leads to the generalizable property of RNR-Map. Also, the decoder can synthesize a novel view, different from the observed direction, based on $m$. 
Examples of reconstructed observations are shown in Figure \ref{fig:recon_random}.
The decoder can render images from novel viewpoints in the observed region, based on the latent codes. 
Note that the rendering boundary is limited to the observed 3D points.
We can see that the decoder generates grey color for the unobserved regions, in the last row of Figure \ref{fig:recon_random}.
More examples of reconstructed images are provided in the supplementary material \ref{appendix:rec_ex}, as well as the network architectures of the encoder and decoder \ref{appendix:network_autoenc}.

\vspace{-0.4cm}
\paragraph{Mapping.}
After training, we can use $F_\mathrm{map}$ for logging visual information from the environment.
Note that the rendering function $F_\mathrm{dec}$ is not needed for mapping.
The \proposed is built incrementally during navigation, based on the odometry information and the known camera intrinsics.
At time $t$, the agent obtains $m_t$ and $n_t$ using $F_\mathrm{map}$, as formulated in (\ref{eq:F_map}).
If the same 3D point in the environment is observed multiple times, $F_\mathrm{map}$ averages the latent codes based on the number of observation instances.
%
% To find out whether averaging duplicated observations is safe for maintaining environmental information, we experimented our model with a trajectory that passes the same position several times. 
% %
% The experiment result is presented in the supplementary material \ref{appendix:averaging latent feature}. 
% %

\section{Localization} \label{sec:loc_framework}
%The objective of building the latent map is to leverage the observed latent codes for the image-goal navigation task.
%
\vspace{-0.1cm}
One of the crucial components of the navigation task is localization.
In the following sections, we describe how \proposed is used for two localization tasks: \textit{image-based localization} and \textit{camera tracking}.
Here, we consider 3-DoF pose $(x, y, a)$ for localization.
%
%In the following sections, we describe the role of \proposed for each navigation task. 
\vspace{-0.1cm}
\subsection{Image-Based Localization}

The implicit visual information of the latent codes in \proposed can provide useful clues for finding out which grid cell is the closest to the given target image $I_\mathrm{trg}$.
Inspired by the fast localization method in \cite{MapNet}, we directly compare latent codes from the target image $I_\mathrm{trg}$ and the \proposed $m$ for localization.
We denote this image-based localization function as $F_\mathrm{loc}$.
Suppose that the target image is taken at the pose $p_\mathrm{trg} = (x_\mathrm{trg}, y_\mathrm{trg}, a_\mathrm{trg})$, and the corresponding map position is $(u_\mathrm{trg}, v_\mathrm{trg})$.
$F_\mathrm{loc}$ outputs a heatmap $E \in\mathbb{R}^{U\times V}$ and the predicted orientation of the target $a_\mathrm{trg}$.
$E$ highlights which grid cell corresponds to the target location $(u_\mathrm{trg}, v_\mathrm{trg})$, among the observed area in $m$.
% %
% If the target location is not observed yet, $E$ can predict the closest area to the target location.
% %

\begin{figure}[t]
  \centering
  \includegraphics[width=\columnwidth, clip, trim=7.0cm 6.5cm 7.0cm 5.1cm]{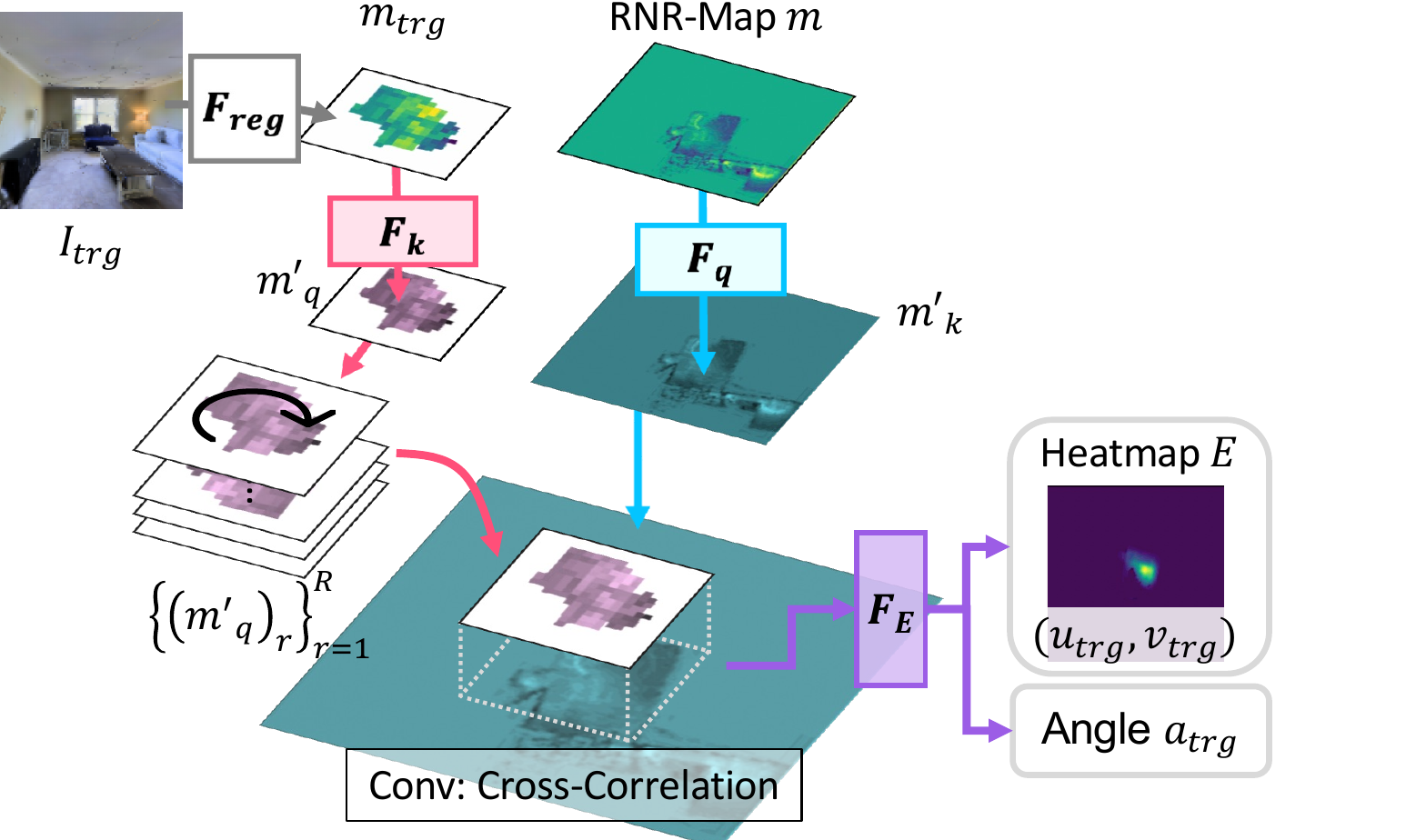}
  
  \caption{\textbf{Localization using $\mathbf{F_\mathrm{loc}}$}. A target observation can be localized by cross-correlation ($\mathrm{Conv}$) between $m$ and $m_\mathrm{trg}$. Before the cross-correlation, each RNR-Map is forwarded to the CNN $F_k$ and $F_q$. After $\mathrm{Conv}$, $F_E$ takes the output of the $\mathrm{Conv}$ and outputs a heatmap $E$ which highlights the most plausible target area.}
  \vspace{-0.2cm}
\label{fig:loc_overview}
\end{figure}

The localization process $F_\mathrm{loc}$ is shown in Figure \ref{fig:loc_overview}.
The \proposed $m$ is constructed from a partial observation about the environment.
The query image is transformed into $m_\mathrm{trg}$ using $F_\mathrm{reg}(I_\mathrm{trg}, p_0 ; \theta_\mathrm{enc}).$
%
% \begin{equation} \label{eq:target_embed}
% \begin{split}
  %  
% \end{split}
% \end{equation}
%
Here, we use origin $p_0 = (0,0,0)$ as an input to $F_\mathrm{reg}$ since the agent does not know the position of the target location.
$F_\mathrm{loc}$ includes three convolutional neural networks, $F_k$, $F_q$, and $F_E$.
$F_k$ and $F_q$ are for proccessing $m$ and $m_\mathrm{trg}$ into $m'_k$ and $m'_q$, respectively.
We found it helpful to introduce neural networks ($F_k$, $F_q$) for better localization results. 
Then, we search query $m'_q$ by convolving (cross-correlation, denoted as $\mathrm{Conv}$) with $m'_k$.
%
%This searching method is adopted from prior approaches that use 2D environmental representation \cite{MapNet, ISS, transporter}.
%
The query map $m'_q$ is rotated into $R$ different angles $\{0^\circ, ..., 360^\circ \times \frac{R-1}{R}\}$. 
$(m'_{q})_r$ denotes $Rot_r(m'_q)$, where $Rot_r$ represents the $r$-th from the $R$ possible rotations.
Each rotated query $(m'_{q})_r$ works as a filter, and the output from the $\mathrm{Conv}$ is forwarded to $F_E$. %
$F_E$ outputs $E \in\mathbb{R}^{U\times V}$ which highlights the most query-related region in $m$.
Each pixel  $e_{u,v} \in E$ in the heatmap represents the probability of the query image being taken at $(u,v)$. 
Also, $F_E$ has an additional head which predict the orientation of the target observation.

The overall localization process can be put together as:
\begin{equation} \label{eq:target_embed}
\small
\begin{split}
    F_\mathrm{loc}(m, m_{\mathrm{trg}}; \theta_{loc}) = F_E(\mathrm{Conv}(F_k(m), \{F_q(m_{\mathrm{trg}})_r\}_1^R)),
\end{split}
\end{equation}
where $\theta_{loc}$ denotes the weight parameters of $F_k$, $F_q$, and $F_E$.
Detailed explanations about the network architecture are provided in the supplementary material \ref{appendix:impd:F_loc}.
We make a ground-truth Gaussian heatmap $E_\mathrm{gt}$ which highlights the ground-truth map location $(u_\mathrm{trg}, v_\mathrm{trg})$ of the query image. 
The representation of an orientation $a_\mathrm{trg}$ is discretized into 18 bins, and $F_\mathrm{E}$ is trained to select the appropriate bin using cross-entropy loss.
The framework is trained to generate a distribution similar to the ground-truth heatmap $E_\mathrm{gt}$ and predicts the $a_\mathrm{trg}$.
The following loss term is used to train $F_\mathrm{loc}$:
\begin{equation}\label{eq:loc}
\begin{split}
    (\hat{E}, \hat{a}_\mathrm{trg}) &= F_\mathrm{loc}(m, m_\mathrm{trg} ; \theta_{loc}) \\
    \mathrm{Loss}(\theta_\mathrm{loc}) &= D_{KL}(E_\mathrm{gt},\hat{E}) + CE(a_\mathrm{trg},\hat{a}_\mathrm{trg}),  \\
\end{split}
\end{equation}
where $D_{KL}$ refers to KL divergence, and $CE$ refers to cross-entropy loss. 
We provide quantitative and qualitative experimental results of $F_\mathrm{loc}$ in the supplementary material \ref{appendix:exp_loc:loc}. 

% \begin{table*}[]
% \begin{tabular}{@{}ccccccc@{}}
% \toprule
% \multicolumn{1}{l}{} & Level 0           & Level 1                          & Level 2                           & Level 3                            & Level 4                            & Level 5                             \\ \midrule
% Img. difference      & \textless{}=  0\% & 0\% \textless{}= x \textless 5\% & 5\% \textless{}= x \textless 10\% & 10\% \textless{}= x \textless 15\% & 15\% \textless{}= x \textless 20\% & 20\% \textless{}= x \textless 100\% \\
% 25cm Recall (\%)     & 77.6              & 75.2                             & 76.6                              & 76.8                               & 76.8                               & 75.9                                \\
% 50cm Recall (\%)     & 99.4              & 99.3                             & 99.3                              & 98.6                               & 100                                & 100                                 \\
% 1m Recall (\%)       & 99.5              & 99.5                             & 99.4                              & 99.3                               & 100                                & 100                                 \\ \bottomrule
% \end{tabular}
% \end{table*}

\subsection{Camera Tracking}
During the navigation, the agent needs to be aware of its own pose to accurately record the observed visual observation.
By cumulating odometry readings, the agent can calculate its relative pose to the start pose.
However, in the real world, it is difficult to determine the accurate pose of a robot due to noises in odometry readings.
The differential rendering function of $F_\mathrm{dec}$ can be used for adjusting the rough estimation of the agent pose $p_t$.
The pose optimization is based on the photometric loss between the rendered image and the current observation. 
As the rendering process $F_\mathrm{dec}$ is differential, the pose of the agent $p_t$ can be optimized with the gradient descent method.
We name this camera tracking function as $F_\mathrm{track}$.
At time $t$, the agent has the previously estimated pose $\hat{p}_{t-1}$, and takes the odometry data $\Delta p_t$.
The rough estimation of the current pose $p_t$ can be calculated by simply adding $\Delta p_t$ to the previous pose $\hat{p}_{t-1}$.
$\bar{p}_t$ denotes such roughly estimated pose: $\bar{p}_t = \hat{p}_{t-1} + \Delta p_t $.
Using $F_\mathrm{track}$, $\bar{p}_t$ is optimized to $\hat{p}_t$.
The output of pose optimization can be derived by the following equation: 
\begin{equation} \label{eq:fine_loc}
\hat{p_t} = F_\mathrm{track}(m_{t-1}, \bar{p}_t) =  \arg\min_{\delta p_t }|F_\mathrm{dec}(m_{t-1}, \bar{p}_t + \delta p_t) - I_t|,
\end{equation}
which minimizes the error between the current observation and the rendered image from the latent map.
$\bar{p}_t$ is the initial value of the pose in the optimization process.
By sampling a small subset of pixels from the image, we can make this optimization process fast enough to use it in navigation.
We provide the accuracy and inference speed of the proposed camera tracking method in the supplementary material \ref{appendix:exp_loc:track}.

\section{Navigation}
\begin{figure}
  \centering
  \includegraphics[width=\linewidth, clip, trim=4.5cm 3.9cm 4.9cm 4.8cm]{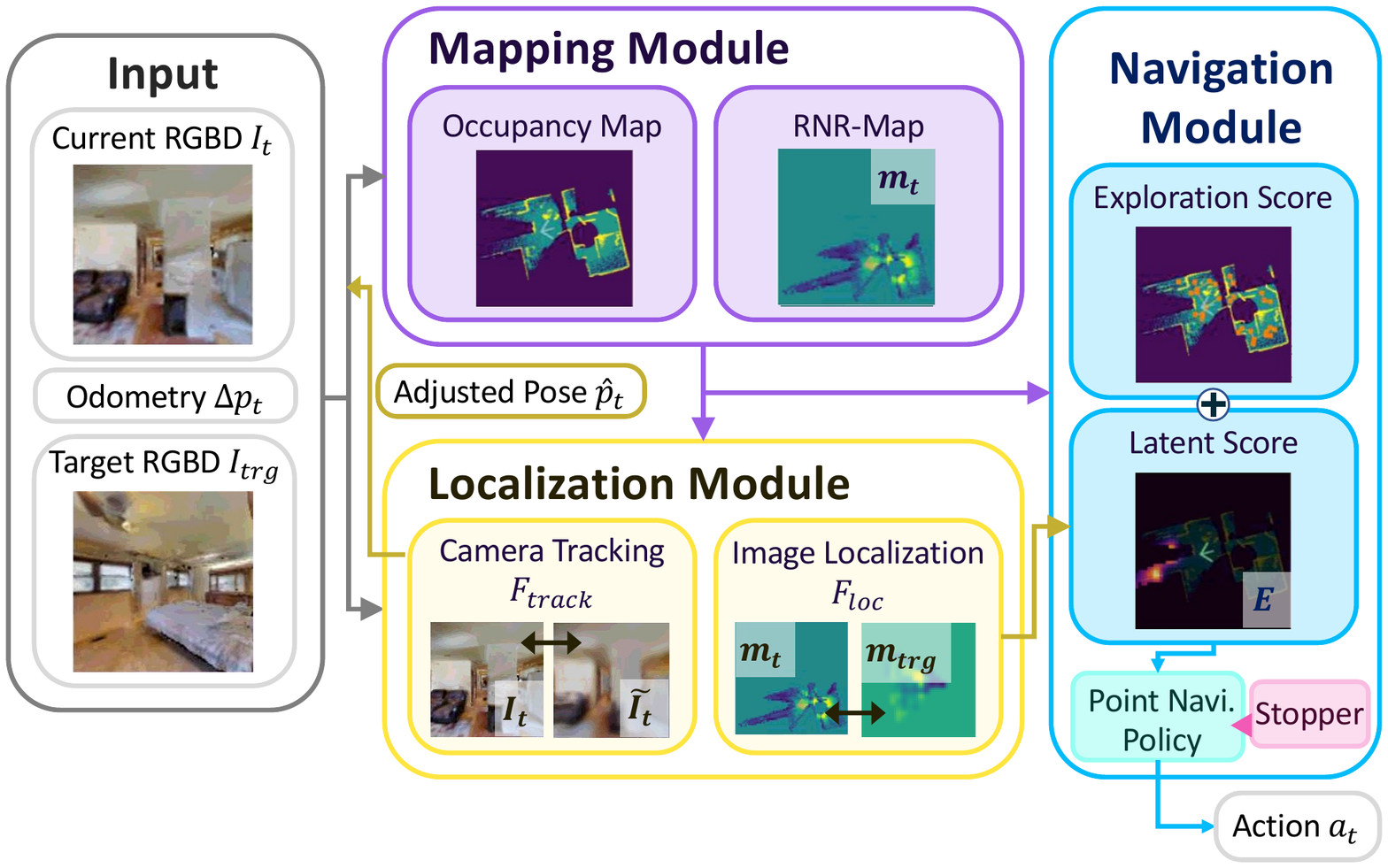}
    \vspace{-0.6cm}
  \caption{\textbf{Navigation System Overview.}}
  \vspace{-0.3cm}
\label{fig:navi_overview}
\end{figure}

\vspace{-0.2cm}
Now we describe how the \proposed and \proposed-based localization functions can be utilized in a navigation system.
We consider the visual navigation task, especially image-goal navigation.
The objective of image-goal navigation is to find the target location given the image taken from the target location.
We build a visual navigation framework which includes $F_\mathrm{map}$ for mapping, $F_\mathrm{track}$ for localization, and $F_\mathrm{loc}$ for target searching.
Figure \ref{fig:navi_overview} shows an overview of the proposed navigation system. 
The system consists of three modules, \textit{mapping, localization}, and \textit{navigation}.
In the following section, we describe how each module works during navigation.  

\subsection{Mapping Module}
The mapping module builds \proposed using the pretrained encoder and $F_\mathrm{map}$. 
At the start of the episode ($t=0$), the mapping module transforms the target image $I_\mathrm{trg}$ into an \proposed  $m_\mathrm{trg}$. 
While maintaining $m_\mathrm{trg}$, the module updates the current \proposed  $m_t$ using each image observation $I_t$ with $F_\mathrm{map}$.
Also, the mapping module builds an occupancy map using depth information.
This occupancy map is used for collision avoidance in a point navigation policy.
The navigation module also uses this occupancy map to add more exploration property to the navigation system.

\subsection{Localization Module}
The localization framework described in Section \ref{sec:loc_framework} works as a localization module in the proposed navigation system.
This localization module has two objectives during navigation.
First, the localization module finds the most probable area which is similar to the target location.
Second, considering a noisy odometry sensor, the localization module is needed to figure out the precise current pose of the agent.
The image-based localization function $F_\mathrm{loc}$ and camera tracking function $F_\mathrm{track}$ are actively used in this module.
With high probability, the target location may not be in the current \proposed $m_t$ in the navigation scenario.
Hence, $F_\mathrm{loc}$ for the navigation task is trained to find the region in $m_t$ which is closer to the target location. 
% %
% With high probability, the target location may not be in the current \proposed $m_t$. 
% Hence, $F_\mathrm{loc}$ for the navigation task is trained to find the closest position in euclidean distance, from the target place among the observed area.
% %
% %We find that the $F_\mathrm{loc}$ trained with unseen queries is more helpful than the one trained only with seen queries.
% %
% %
% The camera tracking $F_\mathrm{track}$ is done when adjusting the current pose of the agent.
% %
% For efficiency, we sample $M$ pixels from the current observation $I_t$ and optimize the pose of the agent using (\ref{eq:fine_loc}).
% %
% The adjusted pose is then used in the mapping module.

\subsection{Navigation Module}
The navigation module consists of three submodules: exploration, point navigation, and stopper.
\vspace{-0.4cm}
\paragraph{Exploration.}
The objective of the exploration module is to select the most plausible region to explore.
The module decides where to visit in order to search the target location, based on the probabilty heatmap $E$ from $F_\mathrm{loc}$ in the localization module.
% %
% A simple approach is to select the grid cell which has the highest value in $E$.
% %
% However, when multiple regions show high probability, the module needs to plan in what order to visit the candidate area. 
% %
% Since the \proposed  and the heatmap $U$ have a dense grid form, simply sampling the grid cells of high probability would not result in an efficient exploration path.
% %
We have adopted the concept from robot exploration \cite{gvg_10,gvg_14,gvg_20,gvg_99}, which builds a generalized Voronoi graph on the occupancy map.
We draw a Voronoi graph on the occupancy map and calculate visitation priority scores for each node of the created graph.
Based on the scores, the exploration module selects the nodes to explore.
Two types of scores are used for selecting exploration candidates, the latent score and the exploration score.
The latent score is based on the heatmap $E$ and represents how probable the node is near the target location.
%
%The exploration score is based on the occupancy map, and represents how much of the unseen area will be revealed when the agent navigates to the node.
The exploration score of each node is simply calculated based on the values in the occupancy map.
The occupancy map has three types of value: occupied, free and unseen.
The exploration score of a node is proportional to the number of unseen pixels in the neighborhood of a node.
The visitation priority of a node is determined based on the sum of the latent score and the exploration score.
\vspace{-0.3cm}
\paragraph{Point navigation policy and Stopper.}
The point navigation policy is a simple occupancy-map-based path-following algorithm.
When the exploration target node is selected, the point navigation module draws the shortest path to the target position.
Following the path, the point navigation policy heuristically avoids obstacles using the occupancy map. 
The stopper module determines the arrival at the target location and calculates the relative pose from the target location.
We employ a neural network $F_\mathrm{stop}$ which decides whether the agent is near the target location.
This neural network is trained to output a binary value (1 if the target location is reached and 0, otherwise) based on the $m_{trg}$ and $m_t$.
For efficient target reaching, we adopted the recent last-mile navigation method\cite{SLING} in stopper module.
Based on keypoint matching, the relative pose between the target location and the current location is calculated using Perspective-n-Point \cite{epnp} and RANSAC \cite{ransac}.
After $F_\mathrm{stop}$ detects the target location, the point navigation policy navigates to the target using the estimated relative pose. 
We provide detailed explanations about graph generation and exploration planning in the supplementary material \ref{appendix:impd_navi}.

\subsection{Implementation Details}
The modules that require training are the encoder and decoder, and the neural networks used in $F_\mathrm{loc}$ and $F_\mathrm{stop}$.
We trained them using the same dataset. 
We have collected 200 random navigation trajectories from each scene in 72 Gibson \cite{xiazamirhe2018gibsonenv} environments with the Habitat simulator \cite{habitat}.
The pair of encoder and decoder is first trained, then $F_\mathrm{loc}$ and $F_\mathrm{stop}$ are trained based on the pretrained encoder.
Further implementation details (network architectures and training details) are provided in the supplementary material \ref{appendix:implementation_details}.

% As the agent uses a directional camera, the partial view can be insufficient to verify whether the current location is the target location. 
% %
% To address such issue, we leverage the latent map. 
% %
% The stopper considers two questions: Is the current observation visually close to the target image (\textit{visual closeness}), and is the current local latent map close to the target latent map in latent space (\textit{latent closeness}). 
% %
% We trained a simple siamese network for determining visual closeness. 
% %
% Latent closeness can be represented by the heatmap $U$ in the Localization Module.
% %
% When the current location shows a high value in both visual closeness and latent closeness, the stopper decides to stop the agent.

\section{Experiments}
We have evaluated \proposed in both localization and navigation tasks.
However, as the main objective of the proposed \proposed is visual navigation, we focus on analyzing the experiment results of the navigation task in the main manuscript.
For localization tasks, we summarize the experimental results here and provide detailed information and analyses in the supplementary material \ref{appendix:exp_loc}. 
\subsection{Localization}
\begin{figure}
  \centering
  \includegraphics[width=\linewidth, clip, trim=5.7cm 4cm 5.5cm 3cm ]{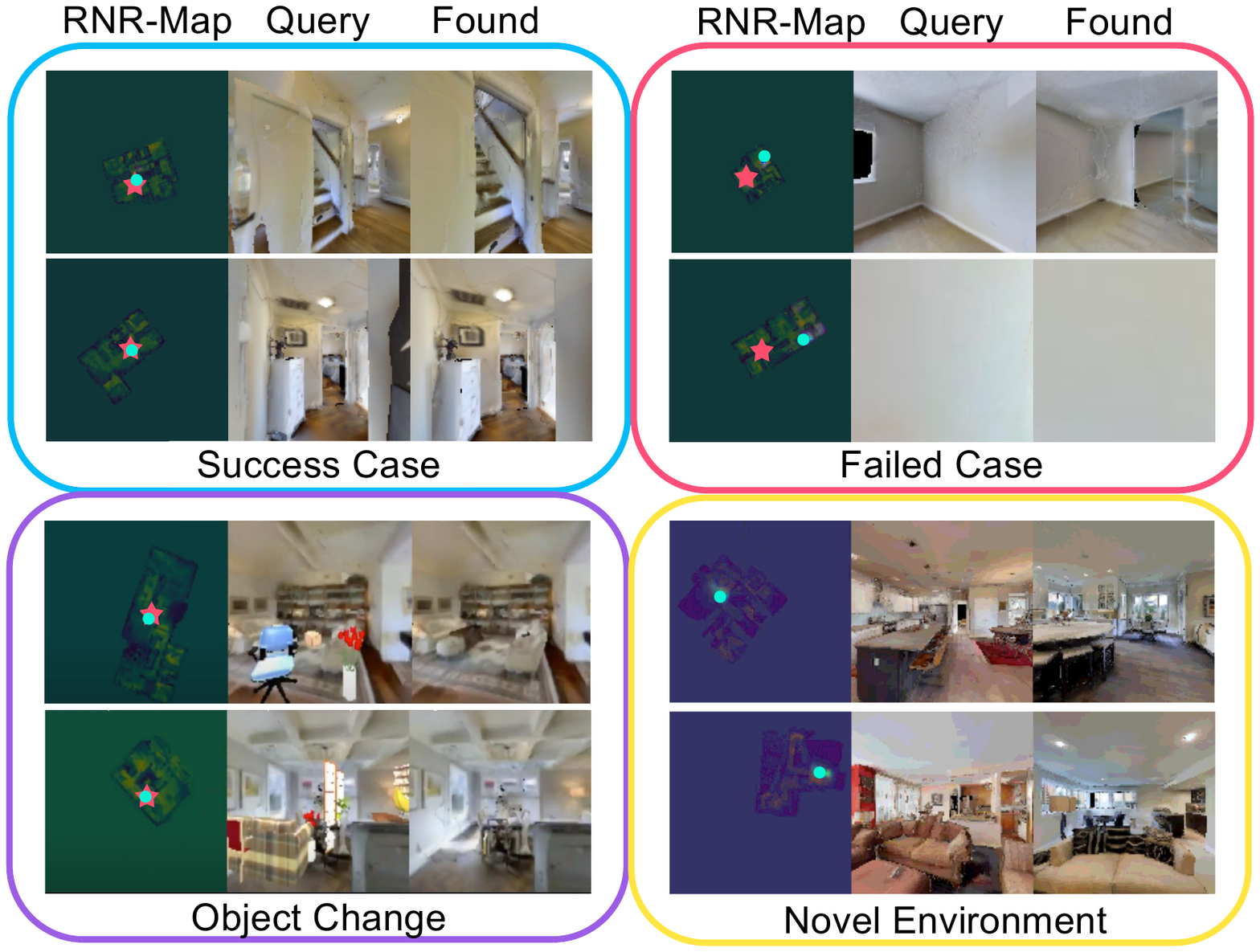}
  \caption{\textbf{Examples of image-based localization}. {\color{RubineRed}\starrr} location of the query image on RNR-Map, and {\color{SeaGreen}\fullcircle} is the location found by  $F_\mathrm{loc}$. More examples are provided in the supplementary material \ref{appendix:exp_loc:loc} and \ref{appendix:exp_loc:sig_loc}.}
  \vspace{-0.3cm}
\label{fig:F_loc_example}
\end{figure}

We have tested $F_\mathrm{track}$ and $F_\mathrm{loc}$ with other related baselines \cite{MapNet, nice-slam}.
In \textbf{camera tracking task}, the \proposed $F_\mathrm{track}$ shows high speed (5Hz) and accuracy (0.108m error) which are adequate for a real-time navigation system.
More than localizing the current pose of the agent, the \proposed $F_\mathrm{loc}$ is able to locate the previously seen images (\textbf{image-based localization task}) with a high recall rate (inliers less than 50cm with 99\%) in the recorded map.
We can leverage this \proposed  for searching the most similar place to the query image even if the exact place is not in the current environment.
Two scenarios can be considered: (1) (Object Change) There have been large changes in the environment so that the object configuration of the environment is different from the information in the \proposed. 
(2) (Novel Environment) The user only has a query image from a different environment but wants to find the most similar place in the current environment.
We have tested the \proposed in both scenarios and observed that $F_\mathrm{loc}$ robustly localizes query images even if some object configuration changes. 
When 33.9\% of the observed images have changed, $F_\mathrm{loc}$ localizes the query image with less than 50cm error in 97.4\% of cases, and 20$^\circ$ error in 97.5\% of cases. 
Also, when given a query image from a different scene, the localized location achieves 94.5\% of visual similarity compared to the best possible visual similarity in the given scene.
The examples of the image-based localization tasks are shown in Figure \ref{fig:F_loc_example}.
We can see that $F_\mathrm{loc}$ finds visually similar places based on the \proposed, even when the environment has changed or given in a novel environment.
Additionally, we found that the suggested method $F_\mathrm{loc}$ mislocates when there are multiple, similar locations in the environment or when the query has poor visual information.
We provide more detailed experiments with baselines (MapNet\cite{MapNet}, NICE-SLAM\cite{nice-slam}) and examples in the supplementary material \ref{appendix:exp_loc}.

\begin{table*}[t]
\centering
\resizebox{0.9\linewidth}{!}{%
\input{tables/img_goal_navi_exp.tex}
}
\vspace{-0.2cm}
\caption{\textbf{Image-goal Navigation Result}. SR: Success Rate. SPL: Success weighted by Path Length.}
\vspace{-0.3cm}
\label{tab:nav}
\end{table*}

\begin{table}[t]
\centering
\resizebox{1.0\linewidth}{!}{%
\input{tables/ablation.tex}
}
\vspace{-0.2cm}
\caption{\textbf{Ablation study}. %
The values are the average of the straight and curved scenarios.}
\vspace{-0.3cm}
\label{tab:nav_ab}
\end{table}

\subsection{Image-Goal Navigation}
%In this section, we explain the baselines we used for the navigation experiments. %The baselines are presented in order based on the theme of the comparison.  
\subsubsection{Baselines}
\vspace{-0.2cm}
We compare \proposed with learning-based agents using behavior cloning (\textbf{BC+RNN}) and reinforcement learning (\textbf{DDPPO} \cite{DDPPO}). Also, to figure out the advantages of the \proposed over an occupancy map, we include the occupancy-map-based coverage method \cite{occ_ans} as a baseline. The objective of this method is to visit all possible area in the given scene.
We modified this method with the target distance prediction from \cite{NRNS}, to make the agent reach the target when it is detected while exploring the environment (\textbf{ANS \cite{occ_ans}+Pose Pred}).
We also compare our method with the recent state-of-the-art image-goal navigation methods.
%(\textbf{NRNS\cite{NRNS}, ZSEL\cite{ZSEL}, OVRL\cite{OVRL}, NRNS+SLING} and \textbf{OVRL+SLING\cite{SLING}}). 
\textbf{ZSEL}\cite{ZSEL}, and \textbf{OVRL}\cite{OVRL} are reinforcement learning-based methods which learn image-goal navigation task with specially designed rewards and the pretrained visual representation.
\textbf{NRNS} builds a topological map and select the nodes to explore by predicting the distances between the target image and the node images with a distance prediction network.
\textbf{SLING} is a last-mile navigation method which predict the relative pose of the target based on keypoint matching, after the target is detected.
This method needs to be integrated with the exploratory algorithm such as NRNS or OVRL. 
Note that our method adopted this method, for efficient target reaching. 
The digits of the baseline DDPPO, OVRL, (NRNS,OVRL)+SLING are from \cite{SLING} and their open-sourced code\footnote{\url{https://github.com/Jbwasse2/SLING}}, and NRNS, ZSEL are from the original papers \cite{NRNS}, \cite{ZSEL}, respectively. 
\vspace{-0.3cm}
\subsubsection{Task setup}
\vspace{-0.2cm}
We have tested each method in the public image-goal navigation datasets from NRNS\cite{NRNS} and Gibson \cite{xiazamirhe2018gibsonenv} with Habitat simulator \cite{habitat}. 
Gibson dataset consists of 72 houses for training split, and 14 houses for the validation split.
NRNS dataset consists of three difficulty levels (easy, medium, hard) with two path types (straight and curved).
Each difficulty level has 1000 episodes for each path type, except for the hard-straight set (806).
The objective of the image-goal navigation is to find the target place given the image of the target location. 
The agent only has access to the current RGBD observations and an odometry reading. 
We consider a noisy setting \cite{occ_ans} where the odometry sensor and the robot actuation include noises as in the real world.
The RGBD image observation comes from a directional camera with $90^\circ$ of HFOV. 
A discrete action space is used in this paper with four types of actions: move forward $0.25m$. turn right $10^\circ$, turn left $10^\circ$, and stop. 
The maximum time step of each episode is set to 500. 
An episode is considered a success when the agent takes a stop action within $1m$ from the target location.
Two evaluation metrics are used: success rate (SR) and success weighted by (normalized inverse) path length (SPL) \cite{SPL}, which represents the efficiency of a navigation path.

\vspace{-0.3cm}
\subsubsection{Image-goal navigation Results}
\paragraph{\proposed helps efficient navigation.}
Table \ref{tab:nav} shows the average SR and SPL of each method.
We can see that the proposed navigation framework with the \proposed shows competitive or higher performance compared to the baselines, on image-goal navigation tasks.
Many of the baselines (DDPPO, ZSEL, OVRL, OVRL+SLING) include reinforcement learning which is sample inefficient and computationally heavy, while having relatively simple representation about the environment. 
In contrast, the \proposed shows higher performances while only requiring an offline trajectory dataset for training neural networks.
Based on this result, we argue that having a good internal representation of the environment and finding how to extract exploratory signals from such representation are important.
An agent with an informative environmental representation can navigate well without the numerous inefficient interactions often required in reinforcement learning.
Compared to baselines which have their own internal representation of the environment (NRNS, NRNS+SLING, ANS), our method shows a much higher performances in curved scenarios.
From this result, we can infer that the \proposed indeed provides useful information for searching targets, more than coverage signals, and better than the existing methods.
The ablation study shown in Table \ref{tab:nav_ab} also displays a similar trend.
We ablated the main functions of the navigation framework $F_\mathrm{loc}$ and $F_\mathrm{track}$, as well as noises.
Without the latent score from $F_\mathrm{loc}$, the success rate and SPL dramatically drop.
We also have tested the proposed method in MP3D \cite{mp3d} dataset, and observed similar results with Gibson dataset. We provide the results in the supplementary material \ref{appendix:navi_mp3d}, and additional ablation study about the submodules of the navigation module is also provided in \ref{appendix:navi_ab}.
\vspace{-3mm}
\paragraph{$\mathbf{F_\mathrm{track}}$ makes \proposed robust to noise.}
Comparing the third row and the last row of Table \ref{tab:nav_ab}, we can infer that the pose adjusting function $F_\mathrm{track}$ helps the agent to find the image goal more effectively, even with the noisy odometry. 
The proposed method shows higher success rates in noisy settings, while SPL values are low.
We believe this is from the randomness of noises, which helps the local navigation policy to escape from being stuck in a single location.

\begin{table}[t]
\centering
\resizebox{0.9\linewidth}{!}{
\begin{tabular}{@{}cccc@{}}
\toprule
Mapping $F_\mathrm{reg}$ & Tracking $F_\mathrm{track}$ & Localization $F_\mathrm{loc}$ & Rendering $F_\mathrm{dec}$\\ \midrule
91.9 Hz (10.9 ms)    & 5.0 Hz (200 ms)   & 56.8 Hz (17.6 ms)        & 14.7 Hz (68.0 ms)     \\ \bottomrule
\end{tabular}}
\vspace{-0.2cm}
\caption{\textbf{Runtime analysis of \proposed.}}
\vspace{-0.5cm}
\label{tab:time}
\end{table}

%\vspace{-3mm}
\paragraph{\proposed is real-time capable.}
We analyzed the runtime for each feature of \proposed and report them in Table \ref{tab:time}.

The runtimes are measured on a desktop PC with a Intel i7-9700KF CPU @ 3.60GHz, and an NVIDIA GeForce RTX 2080 Ti GPU.
We can see that each function of \proposed operates fast enough for real-time navigation, even when including NeRF-based rendering.

\begin{figure}
  \centering
\vspace{-0.1cm}
  \includegraphics[width=\linewidth, clip, trim=4.0cm 6.8cm 4.9cm 4.5cm]{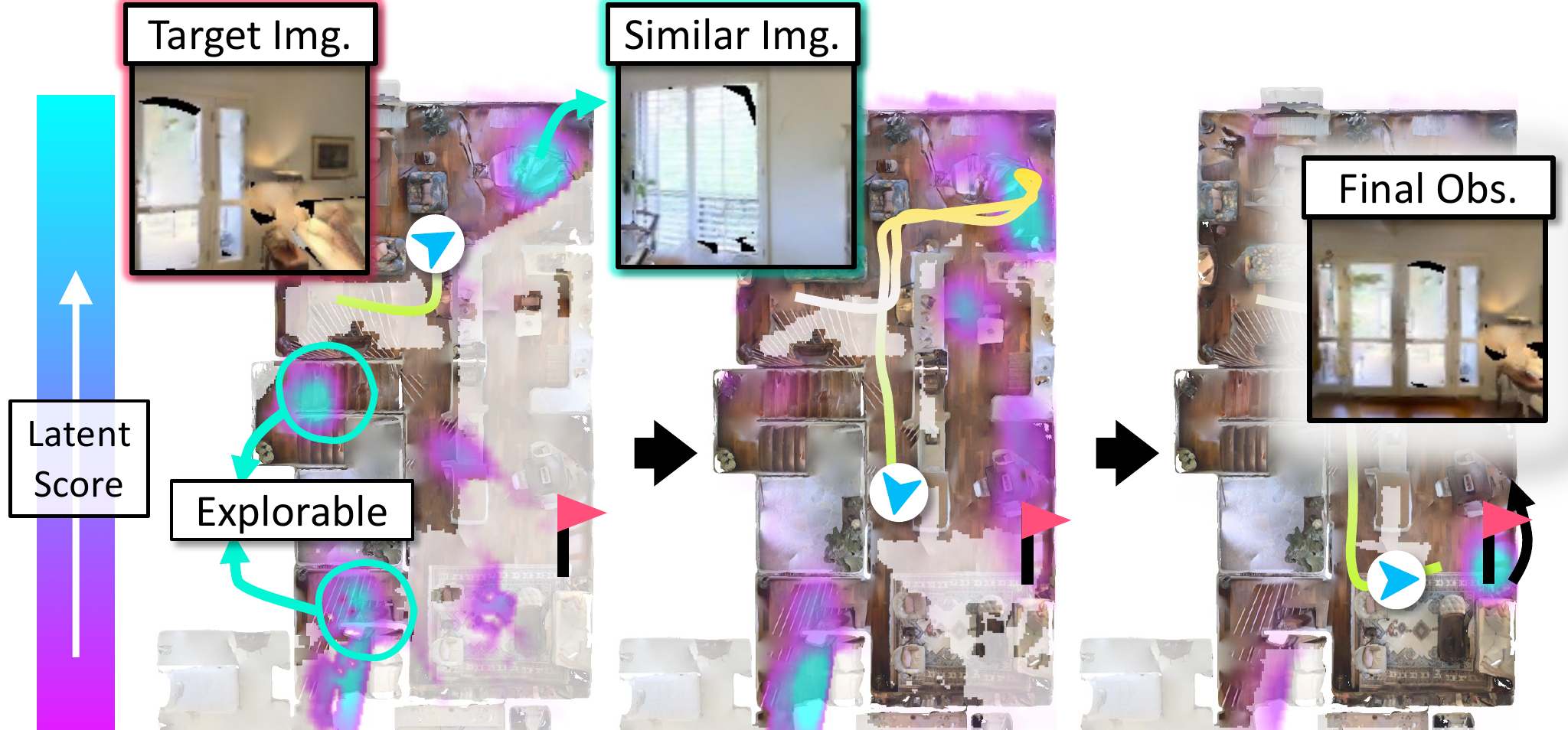}
  \vspace{-0.7cm}
  \caption{\textbf{Example of image-goal navigation episode.} The heatmap value of $E$ from $F_\mathrm{loc}$ is highlighted on the map according to the colorbar on the left.}
  \vspace{-0.0cm}
\label{fig:navi_ex}

\end{figure}

\paragraph{Navigation example.}
Figure \ref{fig:navi_ex} provides an example of an image-goal navigation episode. 
$F_\mathrm{loc}$ highlights two types of places, a place that looks similar to the target image and an explorable place like the end of the aisle, or doorway.
We can see that the highlighted area at the start of the episode has a similar observation to the target image.
First, the agent navigates to a similar-looking place but decided not to stop at the location.
While navigating to another place that shows a high latent score, the agent has found the target location, and the latent score also changed to highlight the target.
Additional qualitative results of navigation episodes are provided in the supplementary material \ref{appendix:navi} and video.
\vspace{-0.1cm}
\section{Conclusion}
\vspace{-0.2cm}
In this paper, we have proposed \proposed  for visual navigation, which captures visual information about the environment.
The \proposed is helpful for visual navigation in two ways: (1) The latent codes in the \proposed can provide rich signals to find the target location, given a query image.
(2) The rendering property of the \proposed helps the agent to accurately estimate its pose based on a photometric error, in an unseen environment.
The proposed method has outperformed other methods in image-based localization.
Also, we have found that the image-based localization of \proposed is robust to environmental changes.
In image-goal navigation tasks, the proposed method outperforms the current state-of-the-art image-goal navigation methods.
Furthermore, the fast inference time of the \proposed shows its potential for real-world applications.
However, the proposed method still has limitations.
Once the observation images are embedded in grids, RNR-Map is hard to correct the past odometry error. 
We can consider applying loop closure using the proposed localization framework. 
Inspired by existing graph-based SLAM methods, a pose graph with local RNR-Maps can be leveraged to optimize poses, leading to consensus in the pixel renderings from the global RNR-Map.
%

% \section{Acknowledgement}
% This work was supported in part by the Institute of Information and Communications Technology Planning and Evaluation under Grant 2019-0-01190, [SW Star Lab] Robot Learning: Efficient, Safe, and Socially-Acceptable Machine Learning, and in part by National Research Foundation under Grant NRF-2022R1A2C2008239, funded by the Korea Government (MSIT). (Corresponding author: Songhwai Oh.)
% The authors are with the Department of Electrical and Computer Engineering and ASRI, Seoul National University, Seoul 08826, Korea.
%%%%%%%%% REFERENCES
{\small
\bibliographystyle{ieee_fullname}
\bibliography{egbib}
}

\clearpage
\appendix

\section*{Appendix: Renderable Neural Radiance Map for Visual Navigation}
We provide additional analyses and examples of the proposed method. The followings are included in this supplementary material:
We provide additional analyses and examples of the proposed method. The followings are included in this supplementary material:
\begin{itemize}
    \item[\ref{appendix:rec_ex}] Examples of reconstructed images from $F_\mathrm{dec}$ 
    \item[\ref{appendix:implementation_details}] Implementation details of the neural networks
    \item[\ref{appendix:exp_loc}] Experiments of localization functions $F_\mathrm{track}$ and $F_\mathrm{loc}$
    \item[\ref{appendix:navi_mp3d}] Image-goal navigation results on MP3D\cite{mp3d} dataset.
    \item[\ref{appendix:navi_ab}] Ablation studies of the modules in the navigation system.
    \item[\ref{appendix:navi}] Qualitative examples of image-goal navigation episodes
    \item[\ref{appendix:impd_navi}] Implementation details of each submodule in the navigation module.
\end{itemize}

\section{Examples of reconstructed images} \label{appendix:rec_ex}
\begin{figure*}[t]
  \centering
  \includegraphics[width=0.85\linewidth, clip, trim=2cm 3cm 2cm 3cm
]{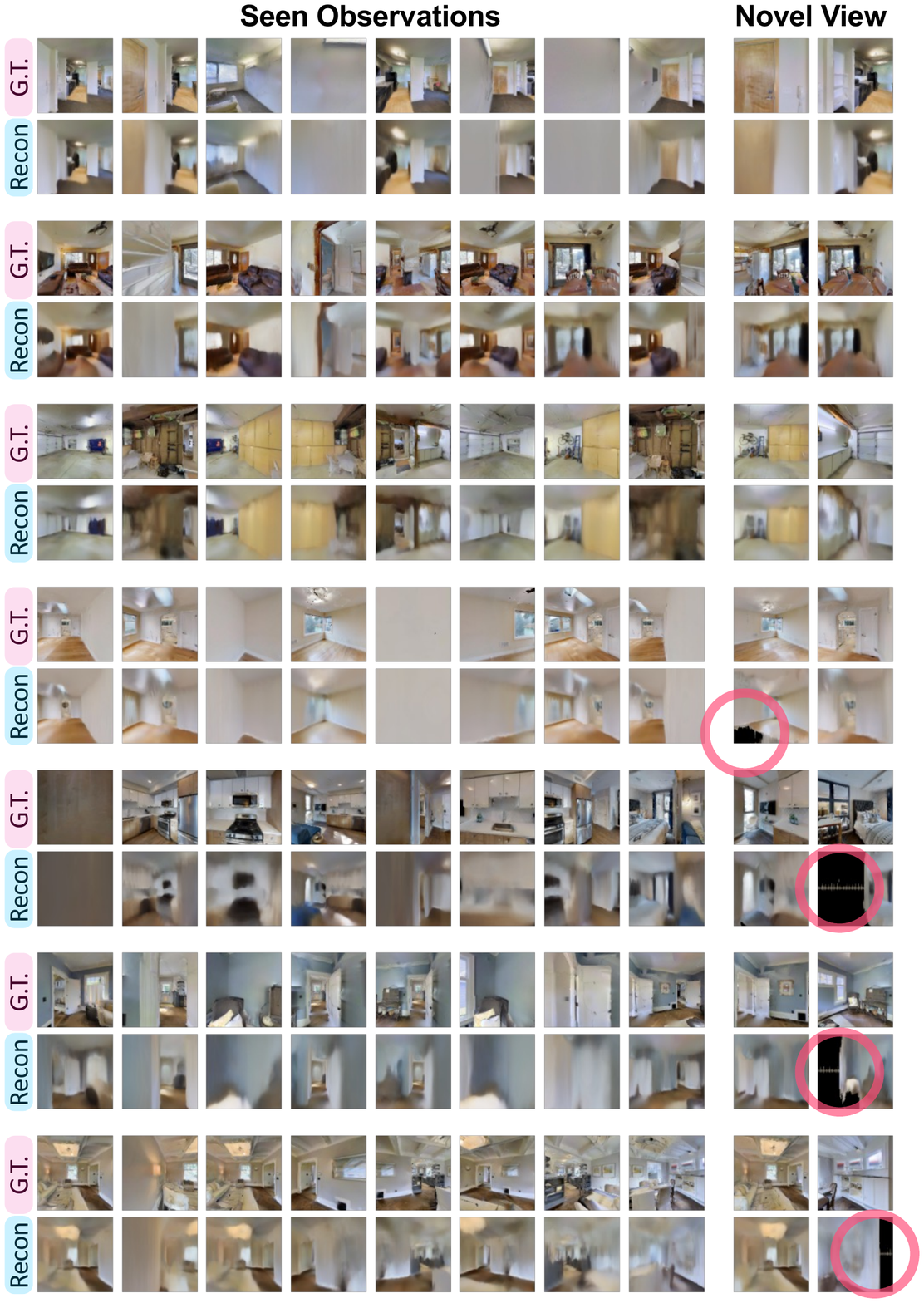}
  \caption{\textbf{Examples of the reconstructed images.} The odd rows are the ground-truth images, and the even rows are the reconstructed images from $F_\mathrm{dec}$. In each row, eight images are embedded in RNR-Map. The last two columns are reconstructed from a novel view. Note that unobserved regions are rendered as an empty space (marked with red circle {\color{RubineRed}\emptycircle}). Each image is rendered with size 128$\times$128.}   
\label{fig:sup:recon_ex}
\end{figure*}

We provide examples of the reconstructed images from the decoder $F_\mathrm{dec}$ in Figure \ref{fig:sup:recon_ex}.
The images are sampled from unseen environments, and not used during training.
We embedded eight image observations for each scenario into RNR-Map and reconstructed the images.
Also, we sample two novel views, which are not embedded in RNR-Map.
We can see that $F_\mathrm{dec}$ can render images well in both seen and novel views.
More importantly, RNR-Map is able to embed and reconstruct images from arbitrary scenes.
RNR-Map can only render the observed region from various viewpoints, and the unobserved regions are rendered as empty space.
Examples of the rendered unobserved region are shown in the last three cases in Figure \ref{fig:sup:recon_ex}.

Although the quality of each image is not state-of-the-art, RNR-Map can reconstruct the overall structure of the image and render large objects.
Small objects and the details of the texture are often ignored in rendered images.
The experiment results from the image-goal navigation task show that it is enough to accurately infer the camera pose based on the image renderings.
Better training techniques and more weight parameters of the encoder and decoder will improve the image quality and navigation performance.

\section{Network Architecture and Training Details}\label{appendix:implementation_details}

\begin{figure}[t]
  \centering
  \includegraphics[width=\linewidth, clip, trim=5cm 5cm 5cm 6cm]{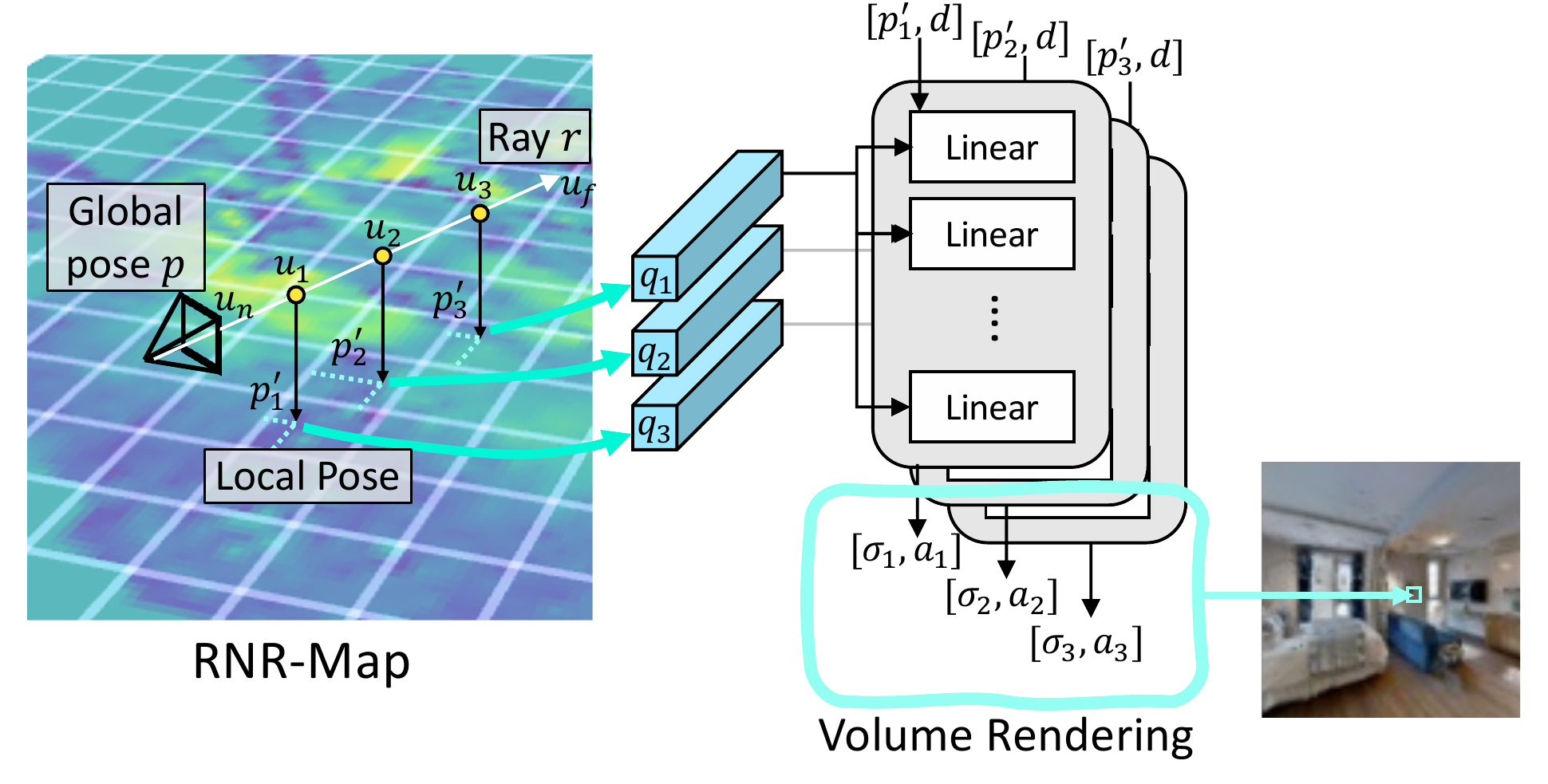}
  \caption{\textbf{Illustration of decoding process.}}
\label{fig:supp:decoder}
\end{figure}

\subsection{Encoder and Decoder}\label{appendix:network_autoenc}
We use images with a size of $128\times128\times4$.
The encoder used in $F_\mathrm{enc}$ is a simple convolutional network consisting of four convolutional layers. 
The encoder embeds an image $I$ into a same-sized feature $C$ with 32 channels.
We normalize each pixel feature $c_{h,w}$ along the feature dimension.
These pixel features are averaged according to their (inverse) projected 3D positions and embedded into RNR-Map $m$.

The network structure of the decoder is adopted from GSN \cite{GSN}.
The illustration of the decoding process is shown in Figure \ref{fig:supp:decoder}.
Given a query pose $\mathbf{p}$ and known camera intrinsic, we can calculate which 3D position will be rendered in a specific pixel $(h,w)$.
More specifically, assuming a ray which passes the pixel $(h,w)$, we sample 3D points along the ray.
Let $u$ be the variable of how far the point is from the camera center along the ray.
For each sampled 3D point, we can calculate its map position, and select the corresponding feature $q$ from the $m$.
As the calculated map position will have continuous values, we sample the feature using bilinear interpolation between the grids.
GSN proposed a local coordinate system, which represents a 3D position of a point with a relative pose in a grid. 
The decoder network takes this local pose $\mathbf{p'}$, view direction $\mathbf{d}$, and the sampled latent code.
This can be understood as the decoder rendering how a specific region would look like, from the local pose and view direction.
The latent code becomes the modulation linear layer, and outputs the occupancy $\sigma$, and appearance $\mathbf{a} \in \mathcal{R}^3$
The pairs of occupancy and appearance are calculated for all the sampled latent codes along the ray.
Finally, the rendered color of a ray $\mathbf{r}$ is calculated with implicit volumetric rendering \cite{Nerf}:

\begin{equation}
\begin{split}
    c(\mathbf{r},m) &= \int_{u_n}^{u_f} Tr(u) \sigma(\mathbf{r}(u),q)\mathbf{a}(r(u), \mathbf{d}, q) du \\
    Tr(u)  &= \mathrm{exp} (-\int_{u_n}^{u} \sigma(\mathbf{r}(u), q) du),
\end{split}
\end{equation}
which is the same as the rendering function in GSN \cite{GSN}.

We only have changed the last step of the rendering in GSN. 
For computational efficiency, GSN renders a small-size of feature map and upsamples it as an image using a convolutional network.
Since the objective of GSN is to generate a realistic scene, a whole image with good quality is needed to evaluate the generated scene.
However, RNR-Map needs to individually render a small subset of pixels for the efficient calculation in the camera tracking function $F_\mathrm{track}$.
Hence, we did not use the last convolutional network in GSN, and replaced it with a simple linear layer.
With this linear layer, we can get the color of the selected pixels, not requiring whole image rendering.

The size of the RNR-map represents $32m\times32m$ area with $128\times128\times32$ size of a tensor, where 32 refers to the feature dimension.
Each grid cell in RNR-map represents $25cm \times 25cm$ of a region.

\subsection{Localization Network $F_{loc}$}\label{appendix:impd:F_loc}
The image-based localization is based on two RNR-Map, $m$ and $m_{trg}$.
$m$ is constructed using partial observations from the environment, and $m_{trg}$ is constructed with the given query image.
Note that the query image is embedded in $m_{trg}$ at the origin pose $p_0$, ($m_{trg} = F_\mathrm{reg}(I_\mathrm{trg}, p_0 ; \theta_\mathrm{enc}).$)
The localization proccess $F_\mathrm{loc}$ is done by convolving the given RNR-Map $m\in\mathcal{R}^{128\times128\times32}$ and the target RNR-Map $m_{trg}^{32\times32\times32}$.
As $m_{trg}$ has only the information of the target image at the origin, most of the grid cells are empty.
Hence, we crop and only use the center of $m_{trg}$, to reduce unnecessary computations.

The localization network $F_\mathrm{loc}$ consists of three convolutional neural networks, $F_k$, $F_q$ and $F_E$.
The last neural network $F_E$ has two heads $F_{E_1}, F_{E_2}$, whose outputs are the heatmap $\hat{E}$ and the predicted orientation of the query pose $\hat{a}_\mathrm{trg}$, respectively.
$F_k$, $F_q$, and $F_{E_1}$ have the same U-Net architecture and output the same-size feature as the input.
$F_{E_2}$ is based on ResNet-18, which processes the output of the cross-correlation into the 18-angle bins. 

\subsection{Stopper $F_{stop}$}\label{appendix:impd:F_stop}
\begin{figure}[t]
  \centering
  \includegraphics[width=\linewidth, clip, trim=7cm 3cm 7cm 3cm]{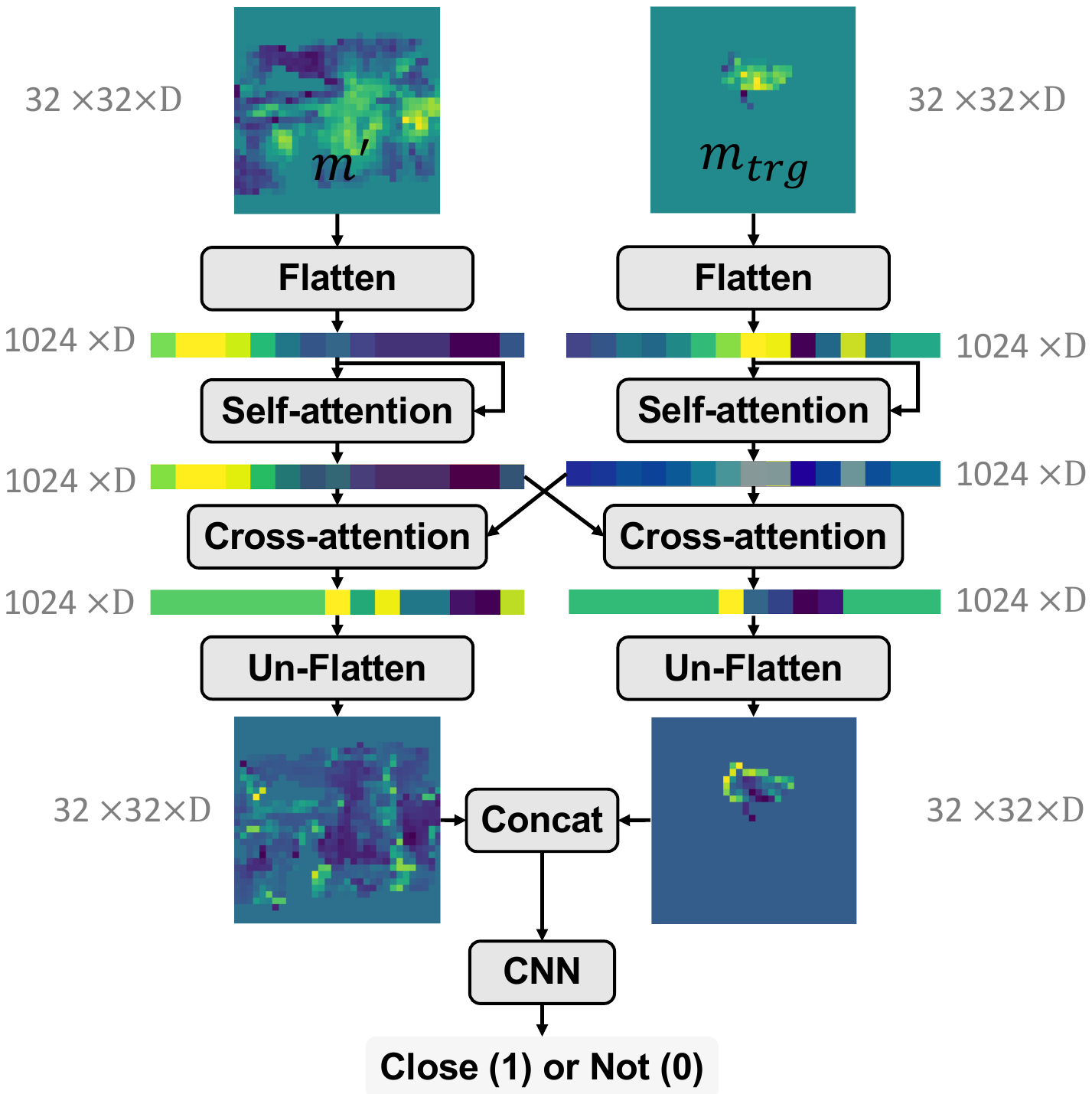}
  \caption{\textbf{Diagram of $\mathbf{F_{stop}}$.}}  
\label{fig:sup:F_stop}
\end{figure}

The stopper $F_\mathrm{stop}$ determines whether the agent is near the target location or not, based on $m$ and $m_\mathrm{trg}$.
%
%We designed this $F_\mathrm{stop}$ to take a closer look of the RNR-Map.
%
We employed the attention mechanism for $F_\mathrm{stop}$, the overall procedure is illustrated in Figure \ref{fig:sup:F_stop}.
The local area has the majority of the information required to determine the relative distance to the target.
Hence we crop the neighborhood of the agent position in $m$, and provide it as an input to $F_{stop}$.
The cropped patch $m'\in\mathcal{R}^{32\times32\times D}$ and the target $m_\mathrm{trg}$ are flattened and considered as a sequence of the latent codes.
We conduct self-attention for each sequence, and then conduct cross-attention between them as shown in Figure \ref{fig:sup:F_stop}.
The outputs of the cross-attention are unflattened as the original size and forwarded to a convolutional network.
This convolutional network consists of four convolutional layers and a linear layer.
The last linear layer outputs the prediction of the closeness to the target.

\subsection{Training}
We have collected 200 random navigation trajectories on 86 (training 72 + validation 14) scenes from the Gibson \cite{xiazamirhe2018gibsonenv} dataset.
We first trained the pair of encoder and decoder, and then trained the $F_\mathrm{loc}$ and $F_\mathrm{stop}$ with the frozen parameters of the encoder.

The neural networks used in RNR-Map are all trained with the same dataset.
The data sample for each neural network is also created in a similar manner.
We sample a trajectory from the dataset and select a subset of frames from the trajectory.
The pose information of the subset is normalized with the first frame of the sampled subset, considering the first frame as the origin.
The images are embedded in RNR-Map according to the normalized pose.
Then we select a query frame from the trajectory.
For the encoder and decoder, the query image is selected in the subset.
They are trained to reconstruct the query image from RNR-Map.
When $F_\mathrm{loc}$ is trained for image-based localization, the query image is also selected from the subset which is embedded in RNR-Map.
In contrast, when $F_\mathrm{loc}$ is trained for navigation, the query image is often selected outside of the subset, which is not provided in RNR-Map.
Using this data sample, $F_\mathrm{loc}$ is trained to estimate which grid cell would be the closest to the target location.
The $F_\mathrm{stop}$ is trained to determine the query position is in the neighborhood of the origin (the first frame of the sampled subset).

The training of the encoder and decoder is done with four GPUs (24GB NVIDIA GeForce RTX 3090), with a batch size of 16. 
The $F_\mathrm{loc}$ and $F_\mathrm{stop}$ are trained with one GPU (24GB NVIDIA GeForce RTX 3090), with a batch size of 32.
All neural networks are trained until the validation loss converges.
% For the encoder and decoder, we sample a trajectory from the dataset and sample a subset of frames from the trajectory, with maximum length 10. 
% %
% Then the encoder embeds all the given observations, and the decoder is trained to reconstruct a random observation among the embedded observations. 
% %
% The data sample for $F_\mathrm{loc}$ is constructed in a similar way of th

\section{Localization Experiments}\label{appendix:exp_loc}

\begin{table*}[h]
\centering
\resizebox{0.9\linewidth}{!}{
\input{tables/loc_exp}}
\caption{\textbf{Localization Results.} \textbf{(a) Camera Tracking.} ATE: Average Trajectory Error. The inference time is the average time of mapping and tracking time for a single frame. \textbf{(b) Image-Based Localization.} $N$-cm Recall refers to the ratio of the cases which the localization error is below $N$-cm. The inference time is the amount of calculation time for a single query, assuming a map is given.}
\label{table:basic_loc}
\end{table*}

\subsection{Camera Tracking}\label{appendix:exp_loc:track}
We compare the proposed method on the camera tracking task against existing methods that build an environmental map in the latent space.
\textbf{MapNet}\cite{MapNet} proposed a method to build an allocentric spatial memory from egocentric observation. 
This method builds a spatial memory for camera pose localization, which is done by comparing the latent features between the current image and the memory.
The localization function $F_\mathrm{loc}$ in our method resembles this operation.
\textbf{NICE-SLAM}\cite{nice-slam} builds a renderable latent map of the environment using a differentiable rendering function (NeRF).
Localization and mapping of this method are based on the photometric error between the rendered image and the observations.
%
%The camera tracking function $F_\mathrm{track}$ in our method resembles this operation.
%First, the pose-localization task can be considered as the camera-tracking task in SLAM methods. 
%
We used the official repository of each method\footnote{MapNet: \url{https://github.com/jotaf98/mapnet}, NICE-SLAM: \url{https://github.com/cvg/NICE-SLAM}}. 
The objective of this task is to estimate the camera trajectory, following a stream of image observations and noisy odometry sensor readings.
We evaluate RNR-Map and MapNet with 1,000 trajectories from the validation scene, and a 10\% subset for NICE-SLAM.
We were only able to test a small subset for NICE-SLAM due to its large computational time.
% {\textbf{\textcolor{blue}{[Is it unfair? :/ But the accuracy of NICE-SLAM is much higher and we are only addressing that NICE-SLAM takes a lot of time. Do I need only to report NICE-SLAM*, not the NICE-SLAM which shows low performance?]}}
Each trajectory contains 1,000 frames of observations.
We use root mean squared error of average trajectory error (ATE RMSE) \cite{ate} as a metric for the accuracy of camera tracking.
The inference time is also measured, and it is the average mapping and tracking time for a single frame. 
The inference times of every method are measured on a desktop PC with a Intel i7-9700KF CPU @ 3.60GHz, and an NVIDIA GeForce RTX 2080 Ti GPU.

The experiment results are shown in Table \ref{table:basic_loc}.
Raw noise in the first row shows the average trajectory error when the noises are not adjusted by any camera-tracking method.
Our method is slightly slower than MapNet.
The reason is that the localization function of MapNet is based on cross-correlation between the latent maps, while our camera tracking function $F_\mathrm{track}$ is based on rendering-based optimization.
However, our method shows higher accuracy in inferring camera poses. 
MapNet discretizes the environment into grids and selects the most relevant grid cell for localization.
In contrast, by rendering the image observations, we can adjust the camera pose at a finer level, smaller than the grid size.

Since our method directly embeds the image feature to the grids, \proposed shows a much faster inference speed because NICE-SLAM needs a rendering-based optimization process for mapping.
The NICE-SLAM camera tracking results are significantly worse than the performances described in the original paper \cite{nice-slam}. 
This is because our trajectory dataset is for navigation.
There is much less overlap between each frame than in the dataset used for SLAM tasks, which results in performance deterioration.
We also tested a different hyperparameter set for NICE-SLAM which conducts more optimization steps on both mapping and tracking (NICE-SLAM*). 
NICE-SLAM* outputs much more accurate results for the camera tracking.
There is a significant trade-off between accuracy and inference time in NICE-SLAM, as more optimization steps lead to high accuracy on camera tracking but take considerable time.

\subsection{Image-Based Localization}\label{appendix:exp_loc:loc}
The objective of the image-based localization task is to find the pose of the query image, which is observed in the distant past.
This task is different from camera tracking, which asks about the current pose of the agent, requiring relatively recent information.
We tested each baseline on the image-based localization, and the results are shown in Table \ref{table:basic_loc}.
We sampled 10 query images from each trajectory used in camera tracking (a total of 10,000 test samples).
Also, in here, we were only able to test 10\% of the samples for NICE-SLAM (1,000 test samples).

The \proposed is aggregated along the trajectory, preserving past information.
As a result, it can be used to locate a target that has been observed in the distant past.
MapNet \cite{MapNet} is focused on finding the camera pose at the moment, so it uses a recurrent neural network (RNN) for a better understanding of sequential observations.
This RNN makes the method less effective in the image-based localization task because old information can be easily lost.

For NICE-SLAM, we render each pose candidate and select the best pose with the lowest photometric difference.
As the rendering process takes a lot of time, we made NICE-SLAM more privileged, providing the possible pose candidates.
The pose candidates are the recorded poses from the mapping process.
NICE-SLAM shows a much longer inference time because the predicted location always has to be compared in the image domain, requiring a rendering process.
In contrast, \proposed can find the location at high speed with high accuracy, by directly comparing the latent codes rather than rendering every position.
$F_\mathrm{loc}$ of RNR-Map locates the target with less than 50cm with an accuracy of 99\%, with a fast speed of 0.018 seconds.
$F_\mathrm{loc}$ selects the most probable grid which corresponds to $25cm \times 25cm$ region. 
This reduces the performance of the 25cm Recall, which requires precise localization finer than the grid size.
However, compared to rendering each image and comparing them to the query, the \proposed still exhibits 5.87\% higher localization performance in 50cm Recall with an incomparably faster time.

\begin{figure}[t]
  \centering
  \includegraphics[width=\columnwidth, clip, trim=7.5cm 3.0cm 7.0cm 3.0cm]{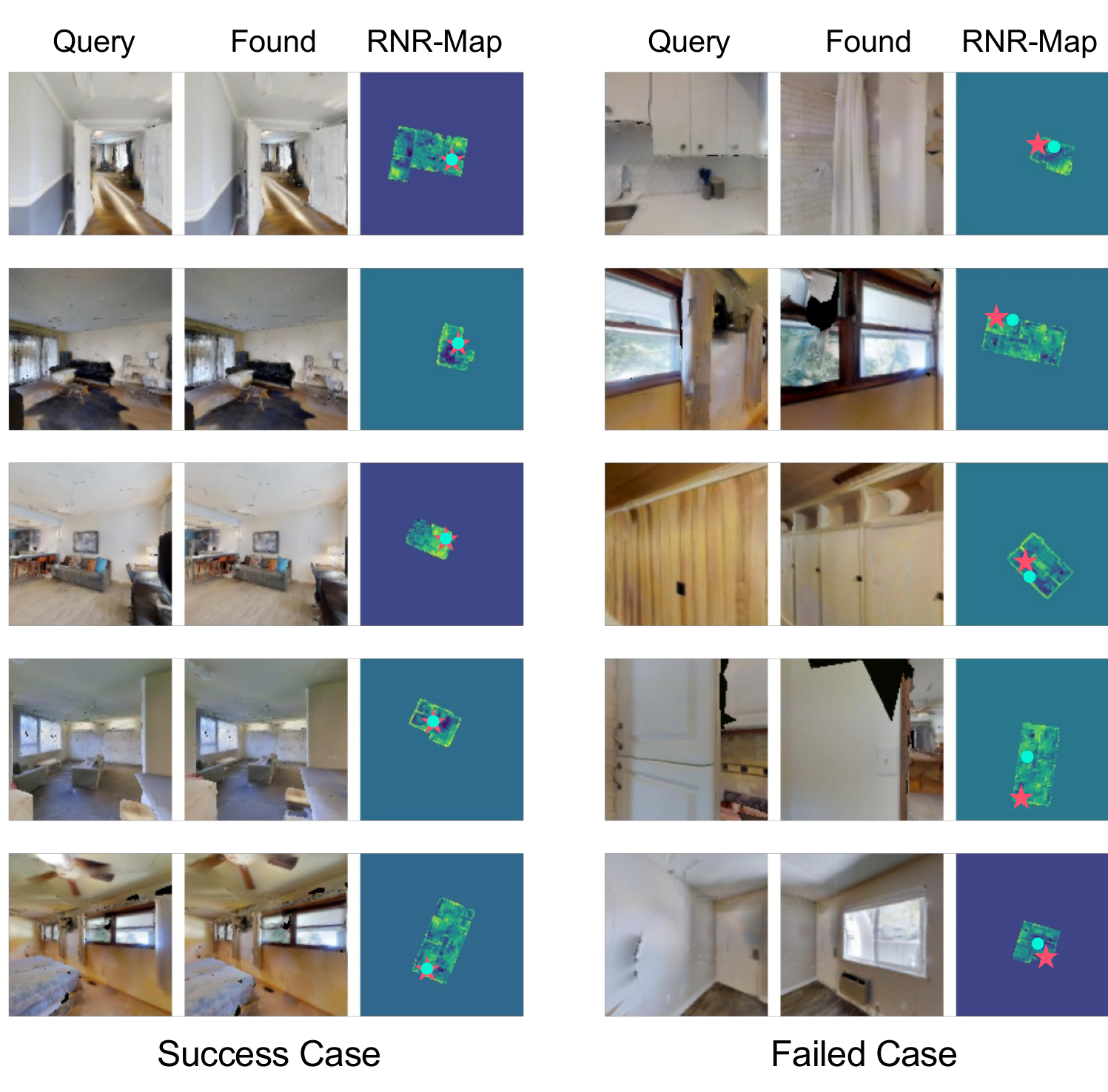}
  \caption{\textbf{Image-based Localization Examples.} The first column shows some examples of successful localization, and the second column shows some examples of failed localization. The query location is marked as red dot, and the predicted localization is marked as a blue dot.}
\label{fig:supp:loc_examples}
\end{figure}

We provide some examples of success cases and failure cases from experiment results in Figure \ref{fig:supp:loc_examples}.
The $F_\mathrm{loc}$ generally finds the query observation with a small error. 
However, when there are multiple visually-similar regions in the given environment, $F_\mathrm{loc}$ often selects the wrong places. 
We can see that the found location has a similar visual appearance to the query image. 

\begin{table}[t]
\centering
\resizebox{0.8\linewidth}{!}{
\input{tables/novel_loc_exp.tex}
}
\caption{\textbf{Localization Results on Object Change scenarios.} Avg Img. Diff.: Average image differences between the query image and the originally observed image. Dist. Err.: Distance error between the localized position and the query position. }\label{tab:ec_loc}
\end{table}

\subsection{Similar-image-goal localization}\label{appendix:exp_loc:sig_loc}
% \label{sec:sig_loc}

We can leverage this \proposed  for searching the most similar place to the query image even if the exact place is not in the current environment.
Two scenarios can be considered: (1) (Object Change) There have been large changes in the environment so the object configuration of the environment is different from the information in the \proposed . 
(2) (Novel Environment) The user only has a query image from a different environment but wants to find the most similar place in the current environment.

\begin{figure*}[t]
\centering
\begin{subfigure}{\textwidth}
  \centering
  \includegraphics[width=0.86\linewidth, clip, trim=1cm 4.5cm 0.8cm 3cm]{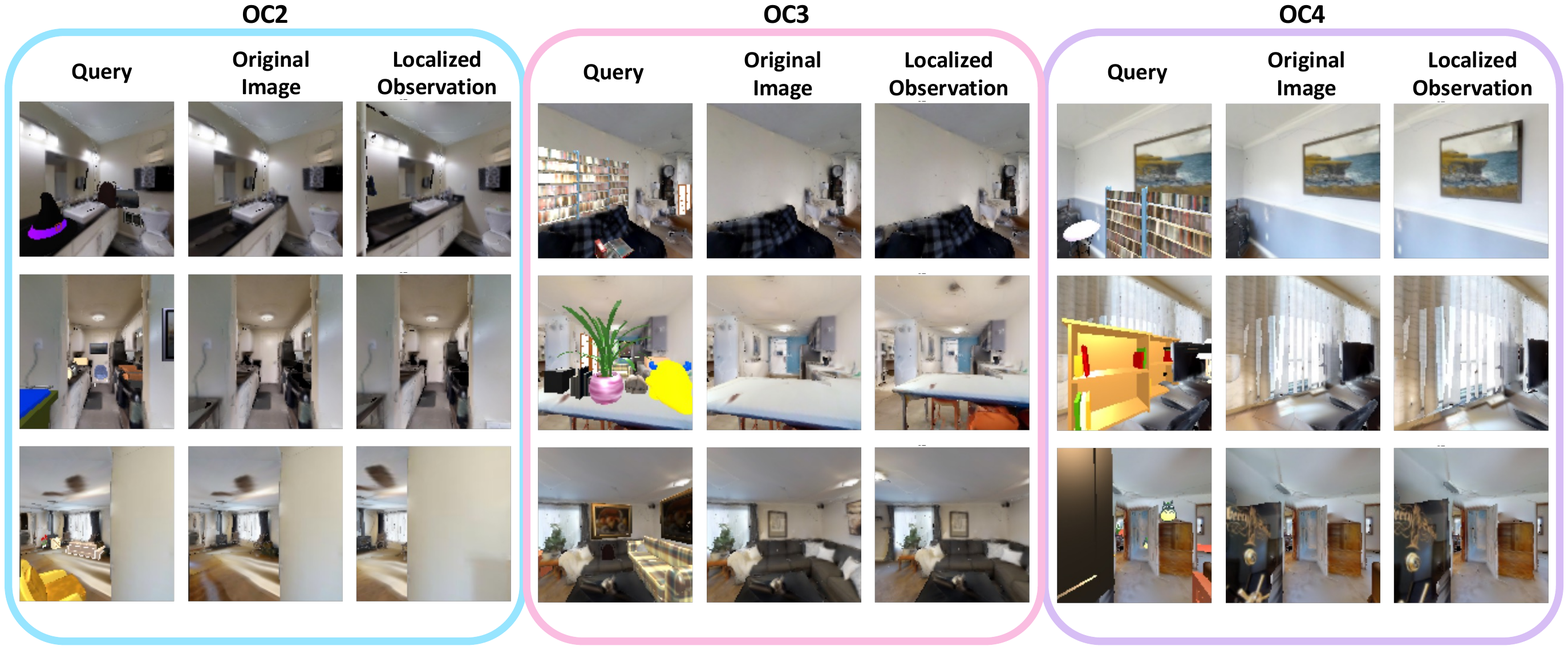}
  \caption{\textbf{Examples of Object Change scenarios.} Examples from OC2, OC3 and OC4 levels are shown in the figure. More examples are provided in the supplementary video.}
\label{fig:supp:oc}
\end{subfigure}
\bigskip
\begin{subfigure}{\textwidth}
\centering
  \includegraphics[width=0.86\linewidth, clip, trim=6.6cm 4.5cm 7.2cm 4cm]{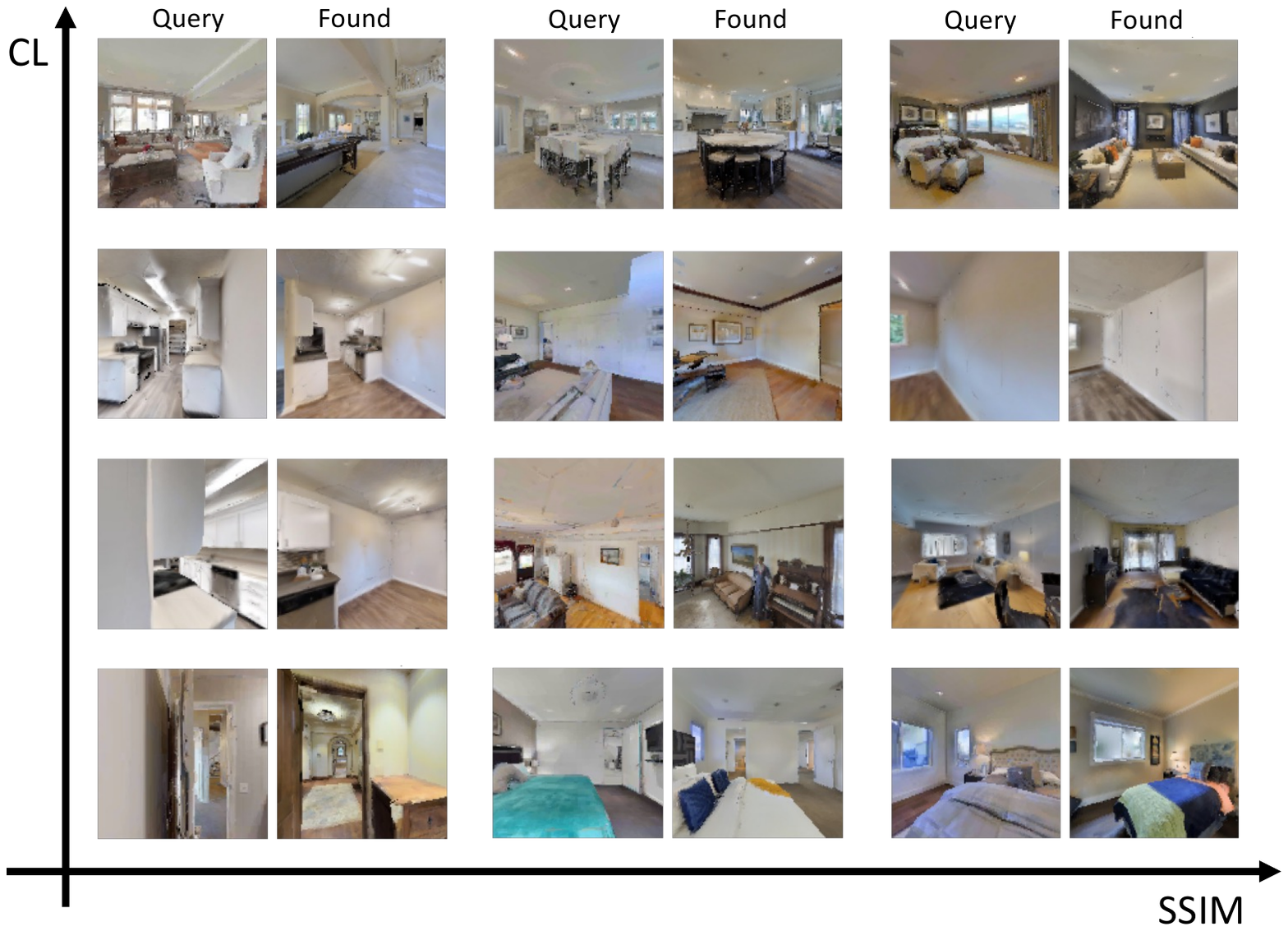}
  \caption{\textbf{Examples of Novel Environment scenarios.} Query is the given localization query from different environments, and Found is the localized observation found by RNR-Map. The image pairs are sorted based on SSIM similarity and CL similarity.}
\label{fig:supp:NE}
\end{subfigure}
\caption{\textbf{Examples of similar-image-goal localization task.}}
\label{fig:supp:SIG}
\end{figure*}

\paragraph{Object change.} 
We divided the difficulty of the localization scenario with the changed environments into four levels: OC1, OC2, OC3, and OC4.
We first recorded \proposed map for each validation scene without any additional objects.
Then we randomly place the random objects (bags, boxes, sofa, chair, toys, \etc) in the scenes, and take pictures of the changed environment.
We gave this picture with objects as a query image.
The difficulty is determined by how much the image has been changed because of the random objects.
The image difference is calculated as the L1 loss between the newly captured image with the random objects and the original image without objects. 
This value is normalized by the size of an image so that it represents how much the image has been changed from the original.
Each difficulty contains 300 samples of the key scene and query image pair.
We show the quantitative results in Table \ref{tab:ec_loc}.
We observed that the \proposed robustly localizes query images even if some object configuration changes.
The \proposed finds the query location with less than 50cm error in 97.4\% of the cases, even in the most difficult scenario.
The qualitative results are shown in Figure \ref{fig:supp:oc}.
%

%
% \begin{figure*}[t]
%   \centering
%   \includegraphics[width=0.8\linewidth, clip, trim=3cm 4.5cm 3.5cm 3cm]{images/supp_OC_ex.pdf}
%   \caption{\textbf{Examples of Object Change scenarios.} Examples from OC3 and OC4 levels are shown in the figure. More examples are provided in the supplementary video.}z
% \label{fig:supp:oc}
% \end{figure*}

\paragraph{Novel environments.} 
We also tested the novel environment (NE), where the query image is from a different scene.
The objective of this task is to find the most visually similar location to the query image.
However, the visual similarity is hard to quantify and can vary depending on the metric.
To evaluate the localized observation, we use various types of metrics which can measure the similarity between images.
We leverage two image encoders which are trained using contrastive learning (CLIP, CL).
In contrastive learning, the image encoder converts images into feature vectors and is trained to maximize the cosine similarity between the features from similar images.
Using the two contrastive learning models, we calculate the cosine similarity of the feature vectors from the query image and the localized observation.
We also utilize the structural similarity index (SSIM) and L1 loss to measure the visual similarity.
The following metrics are employed:
\begin{itemize}
\item \textbf{CLIP} \cite{clip}: CLIP \cite{clip} is a weakly supervised vision-language model with a web-scale dataset, which is trained to embed images and text into a joint latent space. CLIP encoding contains a semantic understanding of an image, and this model is widely used in various kinds of downstream tasks \cite{clip_app_1, clip_app_2, clip_app_3, clip_app_4, clip_app_5, clip_app_6}. We use CLIP for measuring visual similarities between images. The weight parameters from the publicly released model \footnote{\url{https://github.com/openai/CLIP}} are used. This model only takes RGB images.
\item \textbf{CL (SupContrast) \cite{supcontrast}}: We use a contrastive learning model \cite{supcontrast} specially trained on the images from the Gibson \cite{xiazamirhe2018gibsonenv} and MP3D \cite{mp3d} datasets. During the training, similar images are defined by the physical distance between the positions where the image was taken. This model is trained to encode the images from the same region (maximum 2m apart) into similar feature vectors, which show high cosine similarity. This model takes RGBD images. 
\item \textbf{SSIM \cite{ssim} (Structural similarity index)}: We measured SSIM between the depth images of the query image and the localized observation. This measure can represent the similarity between the 3D geometric structures in images.
\item \textbf{(Inverse) L1}: We measure the direct pixel differences between the images. The sum of L1 losses from both RGB and depth images are used. As L1 represents the distance between the images, we inverse the value to use it as a similarity measure. 
\end{itemize}
We tested 100 novel environment scenarios.
As the query image is not from the same scene, the maximum visual similarity may vary depending on the given scene.
We normalize each similarity metric by the maximum possible value from the given scene.
The normalized value represents how much the found location is visually similar compared to the most visually-similar location.
The experiment results are shown in Table \ref{tab:supp:ne_loc}.
The random in the first row shows how much the metrics would appear when we select a random position.

\begin{table}[t]
\centering
\resizebox{0.8\linewidth}{!}{
\input{tables/NE_experiment.tex}
}
\caption{\textbf{Novel Environment Localization Results.} The rows show each search method that finds the most visually similar place based on own metric or method (random, ours). Each column reports the normalized visual similarity from each metric, measuring the image found by each search method in rows.  
Naturally, the measured similarity of a metric found by the same metric would be 100\%. 
}\label{tab:supp:ne_loc}
\end{table}

\begin{figure}[t]
  \centering
  \includegraphics[width=0.8\linewidth]{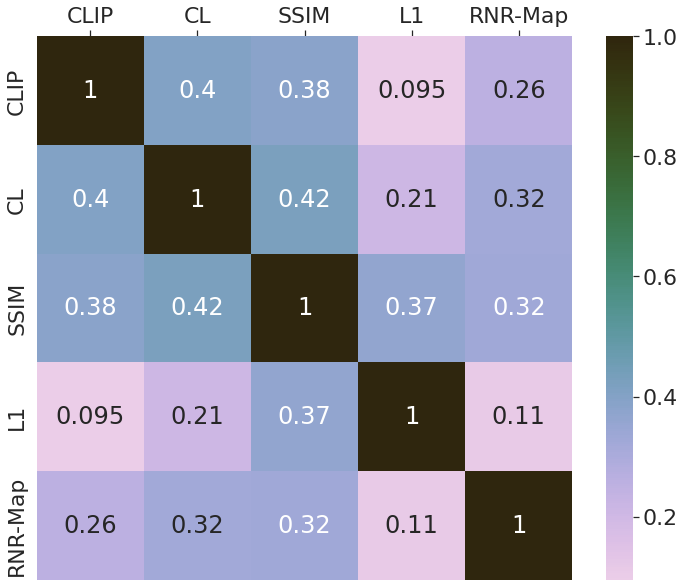}
  \caption{\textbf{Correlation matrix of the similarity metrics and the probability value from RNR-Map.}}% {\textcolor{blue}{\textbf{[I'm not sure whether to include this figure. Is it informative? Is it unnecessary? :/ ]}}}  }
\label{fig:supp:NE_corr}
\end{figure}

\begin{table*}[h]
\centering
\resizebox{0.9\linewidth}{!}{
\input{tables/supp_ablation.tex}}
\caption{\textbf{Ablation studies of the navigation module on Gibson-curved scenarios.} \textbf{Exploration score} and \textbf{Latent score} are used in the exploration module for planning where to visit. \textbf{GT Point Navi} and \textbf{GT Stopper} are ablations of the point navigation policy and the stopper module $F_\mathrm{stop}$, respectively. Each module is replaced with a simple function that has access to ground-truth information from the simulator.}
\label{table:supp:navi_ab}
\end{table*}

We observed that showing high similarity on one metric did not always imply high similarity on other metrics.
We report the experimental results when a query image is searched based on a specific metric (second to the fifth row of Table \ref{tab:supp:ne_loc}).
We sampled all possible locations from the given scene and evaluated each image with the metrics.
For example, the image which shows the highest CLIP visual similarity becomes the localized image by using CLIP as a search method (Max CLIP in Table \ref{tab:supp:ne_loc}).
%
%For example, the second row of Table \ref{tab:supp:ne_loc} represents we selected the localized observation as the image which shows the highest CLIP similarity with the query image. 
%
The CL column of the Max CLIP row shows that the image which has the highest CLIP similarity shows 61.36\% of CL similarity compared to the maximum CL similarity.
We can see that the maximum of each metric does not always mean the maximum in the other similarity metrics.

%The last row of the table shows the localization results from RNR-Map.
%
Meanwhile, the RNR-Map localizes a query image with consistently high value in all metrics (the last row).
The RNR-Map shows general visual similarity in various metrics, in contrast to using a specific metric as a search method.
We can infer that the RNR-Map finds a visually similar location even with a query image from a different environment.

Furthermore, we want to stress that the localization process $F_\mathrm{loc}$ of RNR-Map is done with a forward pass to the neural network with 56.8Hz of speed.
Even with the fast speed, RNR-Map still achieves competitive similarity compared to the cases when all the possible images are compared with the query image one by one.
We also plot the correlation matrix between the metrics and the probability value from RNR-Map in Figure \ref{fig:supp:NE_corr}.
The RNR-Map shows positive correlations with the similarity metrics.
The examples of the query image and the localized observation pairs are shown in Figure \ref{fig:supp:NE}. 
In Figure \ref{fig:supp:NE}, the image pairs are sorted by CL similarity and SSIM similarity.
We observed that SSIM similarity is high when the overall 3D structures of the images are similar, while CL similarity is high when the overall colors of the images are similar.

\begin{table*}[h]
\centering
\resizebox{0.7\linewidth}{!}{
\input{tables/supp_mp3d.tex}
}
\caption{\textbf{Image-goal navigation results on MP3D\cite{mp3d}-curved scenarios in NRNS\cite{NRNS} dataset.} The first column shows the training dataset each method has trained on. }\label{tab:supp:mp3d}
\end{table*}

\section{Navigation ablation studies}
\label{appendix:navi_ab}

In this section, we report the ablation studies of the navigation module.
The results are shown in Table \ref{table:supp:navi_ab}, and each method is evaluated on Gibson-curved scenario from NRNS dataset \cite{NRNS} without noise setting.
\paragraph{Ground-truth information.} 
We provided the ground-truth information to the point navigation policy and the stopper module.
With ground-truth information, the point-navigation policy can reach the target using the shortest path without collision.
The stopper module with the ground-truth information has access to the geodesic distance to the target, so that it can detect the target location and take a stop action accurately. 
We can observe that navigation performance increases when given ground-truth information.
The information about the distance to the target location is critical to the performance.
Both the success rate and SPL dramatically increase because the ground-truth information enables efficient navigation without wandering or collision.
Also, the agent has no risk of taking the false-stop action (terminating the episode even when the target is far) or overlooking the target location.
We observed that the ground-truth point navigation policy often leads to the failure of an episode.
Hence, in the medium scenarios, the model without the ground-truth point navigation policy (full model in the last row) shows better performance than the one with the ground-truth point navigation policy.
This is because of the relatively inefficient path from our point-navigation module, which leads to a random chance to explore the environment more and find the target location.
We anticipate that an improved point navigation policy and better $F_{stop}$ will largely improve the navigation performances.

% {\textcolor{blue}{ \textbf{[The results are quite weird :/. I'm doing the experiments again.]}
% However, in the hard scenario, providing ground-truth information does not yield better navigation performances.
% %
% We believe this is because $F_{stop}$ detects the target location from the relatively far position (more than 1.0m).
% %
% The stopper module with the ground-truth information does not take the stop action unless the distance becomes below 1.0m.
% %
% However, when $F_{stop}$ is used, it detects the target location from a far position and proceeds to the keypoint-matching with the target image.
% %
% The $F_{stop}$ takes a local patch of RNR-Map which is corresponding to $8m \times 8m$ area.
% %
% The stopper module $F_{stop}$ and the keypoint-matching policy rather helps the agent to finds and reach to the target.}}

\paragraph{Exploration strategy.}
The exploration module in the navigation module selects where to explore in order to find the goal location.
The exploration module draws a Voronoi graph on the occupancy map and selects a node as an exploration candidate from the created graph.
Two criteria are used for selecting the exploration candidates, latent score and exploration score.
The latent score is based on the localization heatmap value from $F_{loc}$.
The occupancy map has three types of values, free (1), occupied (2), and unseen (0).
The exploration score is based on the number of unseen pixels in the neighborhood of a pixel.
The more unseen pixels are in the neighborhood, the more likely a location has been underexplored and has a high probability of discovering new areas.
We ablated each score, and the results are shown in Table \ref{table:supp:navi_ab}.
Without the latent score, only the exploration score is used for exploring the environment.
The exploration score helps the agent visit all the possible places in the scene.
However, this score does not consider the information from the target, so the agent may not be able to find the target within the time limitation due to inefficient exploration.
We can see that the agent finds the target more successfully when only the latent score is used.
In the full model, we use the sum of the exploration score and the latent score.
Using the combination of two scores helps the agent to find the target location efficiently.
The latent score can guide the agent to the target location based on the RNR-Map.
However, when there is not much information or clues about the target location, the exploration score can encourage the agent to explore the environment.

\section{Navigation on MP3D Dataset}\label{appendix:navi_mp3d}

We provide the image-goal navigation experiment results on MP3D dataset in Table \ref{tab:supp:mp3d}.
The experiments are done without noise setting, and we report the digits from \cite{NRNS}, \cite{SLING} for the baselines.
The proposed RNR-Map-based navigation framework outperforms the baselines, except for the success rate in easy scenarios.
The neural networks in RNR-Map are trained using Gibson \cite{xiazamirhe2018gibsonenv} dataset. 
Based on the results, we can observe that the RNR-Map-based framework is generalizable to a different dataset, without any fine-tuning.

\begin{figure*}[t]
\centering
\begin{subfigure}{\textwidth}
  \centering
  \includegraphics[width=0.9\linewidth, clip, trim=1cm 4.5cm 0.5cm 4cm]{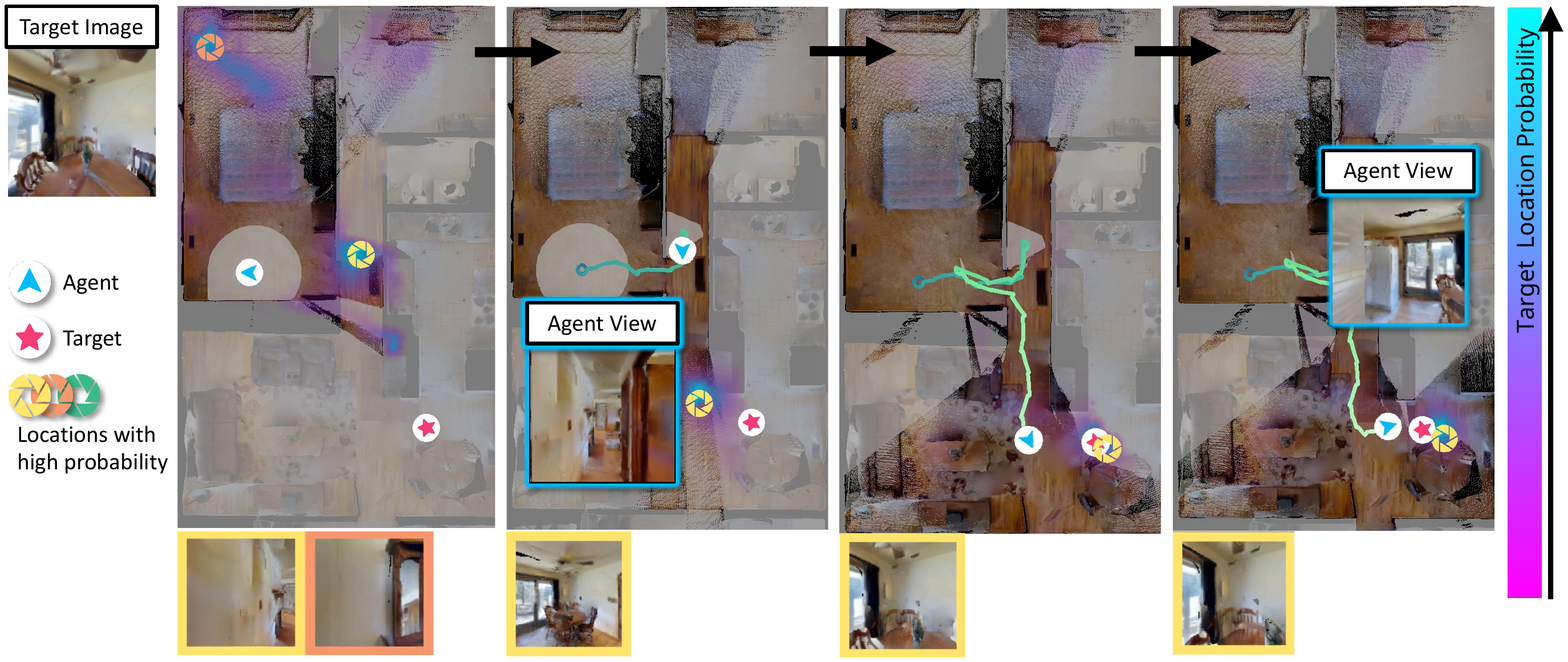}
  \caption{\textbf{Example 1.}}
  \label{fig:supp:navi1}
\end{subfigure}
\bigskip
\begin{subfigure}{\textwidth}
\centering
  \includegraphics[width=0.9\linewidth, clip, trim=1cm 4.5cm 0.5cm 4cm]{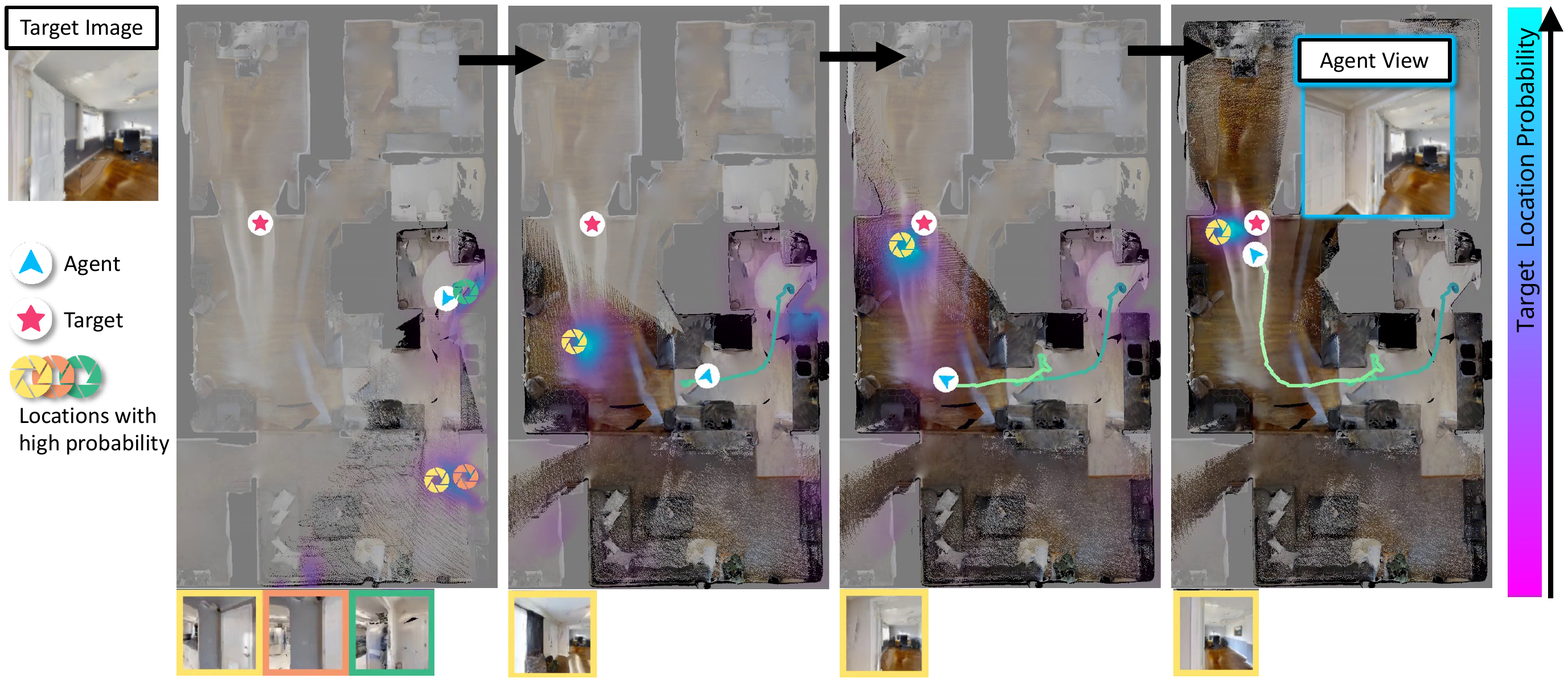}
  \caption{\textbf{Example 2.}}
  \label{fig:supp:navi2}
\end{subfigure}
\caption{\textbf{Examples from image-goal navigation episodes.} The heatmap values (the latent score) from $F_\mathrm{loc}$ are presented on the map according to the color bar on the right. We also marked the locations of high-probability values with the images from the location.} 
\label{fig:supp:navi}
\end{figure*}

\begin{figure*}[t]
  \centering
  \includegraphics[width=0.8\linewidth, clip, trim=1.5cm 7.5cm 1.5cm 5.5cm]{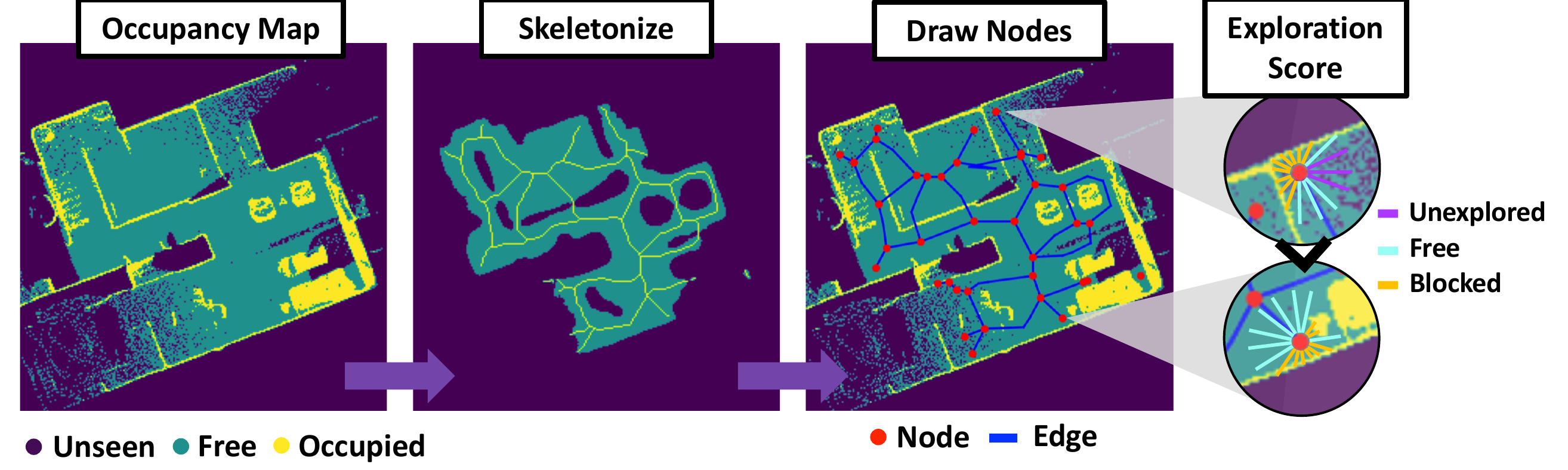}
  \caption{\textbf{Overview of the graph generation.} We take the free area of the occupancy map and skeletonized the image. Then, we draw a graph on the skeleton. Each node has an exploration score, which is calculated based on the number of unseen pixels in the neighborhood.}  
\label{fig:supp:graph_gen}
\end{figure*}

\section{Navigation Examples}\label{appendix:navi}
We provide examples from the image-goal navigation episodes of RNR-Map in Figure \ref{fig:supp:navi}.
The latent scores usually highlight unexplored areas at the beginning of episodes.
This is because the $F_\mathrm{loc}$ for navigation is trained to predict the closer area to the target given the partial information of the environment.
We can see that the way to other rooms is highlighted in the first column of Figure \ref{fig:supp:navi1} and \ref{fig:supp:navi2}.
The agent follows the latent score and expands its RNR-Map according to the image observations.
During the exploration, the agent observed the target-related region and successfully reached the target location.
More examples are provided in the attached video.

\section{Implementation Details for the navigation submodules} \label{appendix:impd_navi}
\paragraph{Graph Generation}
For efficient exploration planning, we discretize the region in the observed environment into a graph.
Figure \ref{fig:supp:graph_gen} shows the process of graph generation.
This method is inspired by robot exploration literature \cite{gvg_10,gvg_14,gvg_20,gvg_99}, which draws an (approximated) Voronoi graph on the occupancy map.
Originally, the Voronoi graph consists of nodes that are equally distanced from the neighbor obstacles.
We simplify this graph construction with the image skeletonizing method in the image processing library (\texttt{skimage.morphology.skeletonize}) \cite{van2014scikit}.
This function is based on the image thinning algorithm proposed in \cite{thinning}.
As the skeleton only consists of lines, we need to determine which pixels will be the nodes.
We select the pixels that have many neighbors as a node.
The neighbor of more than two pixels indicates that the pixel is not on a line, but instead in the intersection of several lines.
Then we split the skeleton based on the selected nodes, and determine the relationship between the nodes.
The created graph is used for planning the exploration of the agent.
The exploration module selects the node to explore, rather than sampling a pixel among the free spaces in the occupancy map.

\paragraph{Exploration Score}\label{appendix:impd_navi:exp}
The exploration module evaluates each node in the graph with two criteria.
The first one is the latent score, which is from the heatmap from $F_{loc}$ in the localization module.
This heatmap highlights the place related to the target image.
The second one is the exploration score, which is based on the number of unseen pixels in the neighborhood of the node.
A location is more likely to have been underexplored and have a high likelihood of discovering new areas if there are more unseen pixels nearby.
The examples of calculating the exploration score are presented in Figure \ref{fig:supp:graph_gen}.
Drawing a set of rays centered at the node, we count the number of unseen pixels in the rays.
We also evaluate whether a ray is blocked by an obstacle, and limit the length of the ray to less than the distance to the blocked obstacle.
The number of unseen pixels is normalized into 0.0 to 1.0, and this value becomes the exploration score.
Furthermore, we also calculate the distance from the agent to each node and add the inverse value to the exploration score.
This encourages greedy exploration, which explores the near neighborhood first.

\paragraph{Point Navigation Module}\label{appendix:impd_navi:point_navi}
The point navigation module takes the map position of the agent and the selected node position to explore. 
This module calculates the collision-free shortest path to the node based on the fast-marching method.
We used the open-source library for the fast-marching method \footnote{\url{https://github.com/scikit-fmm/scikit-fmm}}.
After a navigation path is obtained, the point navigation module outputs an appropriate action based on its relative pose to the path points.

\paragraph{Stopper Module}\label{appendix:impd_navi:stopper}
The stopper module has two components, $F_{stop}$ and the last-mile approaching function which is adopted from \cite{SLING}.
First, $F_{stop}$ determines whether the agent is near the target location.
Second, if $F_{stop}$ found that the target is near the agent, we conduct keypoint-matching \cite{superglue} between the current observation and the target image.
If a sufficient number of keypoints are matched, we can infer the relative pose between the target location and the current position using the depth information.
As proposed in \cite{SLING}, we use Perspective-n-Point \cite{epnp}, and RANSAC \cite{ransac}.
After the relative pose is determined, we set the local goal point, and the point navigation module navigates to the estimated target.
If the number of matched keypoints decreases below a certain threshold (20 in our case), the last-mile approach is terminated, and the exploration module selects the next exploration target.

\end{document}

%% file: tables/img_goal_navi_exp.tex
\begin{tabular}{@{}ccccccccc|ccccccll@{}}
\toprule
\multicolumn{1}{l}{}             & \multicolumn{8}{c|}{\textbf{Stragint}}                                                                                                                                        & \multicolumn{8}{c}{\textbf{Curved}}                                                                                                                                                 \\ \midrule
\multicolumn{1}{l}{}             & \multicolumn{2}{c}{\textbf{Easy}}                   & \multicolumn{2}{c}{\textbf{Medium}}      & \multicolumn{2}{c}{\textbf{Hard}} & \multicolumn{2}{c|}{\textbf{Overall}}    & \multicolumn{2}{c}{\textbf{Easy}} & \multicolumn{2}{c}{\textbf{Medium}} & \multicolumn{2}{c}{\textbf{Hard}} & \multicolumn{2}{c}{\textbf{Overall}}                                  \\
\multicolumn{1}{l}{}             & \textbf{SR}              & \textbf{SPL}             & \textbf{SR}   & \textbf{SPL}             & \textbf{SR}     & \textbf{SPL}    & \textbf{SR}              & \textbf{SPL}  & \textbf{SR}     & \textbf{SPL}    & \textbf{SR}      & \textbf{SPL}     & \textbf{SR}     & \textbf{SPL}    & \multicolumn{1}{c}{\textbf{SR}}   & \multicolumn{1}{c}{\textbf{SPL}}  \\ \midrule
\textbf{BC + RNN}                & 39.4                     & 27.9                     & 25.7          & 15.9                     & 13.3            & 9.0             & \multicolumn{1}{l}{26.1} & 17.6          & 26.4            & 12.6            & 20.3             & 10.5             & 8.4             & 4.8             & 18.4                              & 9.3                               \\
\textbf{DDPPO \cite{DDPPO}}                   & 43.2                     & 38.5                     & 36.4          & 34.8                     & 7.4             & 7.2             & \multicolumn{1}{l}{29.0} & 26.8          & 22.2            & 16.5            & 20.7             & 18.5             & 4.2             & 3.7             & 15.7                              & 12.9                              \\
\textbf{ANS + Target Pred \cite{occ_ans}}       & 68.8                     & 55.1                     & 54.0          & 30.3                     & 42.4            & 22.9            & \multicolumn{1}{l}{55.1} & 36.1          & 48.0            & 21.0            & 46.0             & 20.5             & 31.3            & 14.6            & 41.8                              & 18.7                              \\
\textbf{NRNS \cite{NRNS}}                    & 64.1                     & 55.4                     & 47.9          & 39.5                     & 25.2            & 18.1            & 45.7                     & 37.7          & 27.3            & 10.6            & 23.1             & 10.4             & 10.5            & 5.6             & 20.3                              & 8.8                               \\
\textbf{ZSEL \cite{ZSEL}}                     & -                        & -                        & -             & -                        & -               & -               & -                        & -             & 41.0            & 28.2            & 27.3             & 18.6             & 9.3            & 6.0            & 25.9                              & 17.6                              \\
\textbf{OVRL \cite{OVRL}}                    & 53.6                     & 34.7                     & 48.6          & 33.3                     & 32.5            & 21.9            & \multicolumn{1}{l}{44.9} & 30.0          & 53.6            & 31.8            & 47.6             & 30.2             & 35.6            & 22.0            & 45.6                              & 28.0                              \\
\textbf{NRNS + SLING \cite{SLING}}            & \textbf{85.3}            & \textbf{74.4}            & 66.8          & \textbf{49.3}            & 41.1            & 28.8            & \multicolumn{1}{l}{64.4} & \textbf{50.8} & 58.6            & 16.1            & 47.6             & 16.8             & 24.9            & 10.1            & 43.7                              & 14.3                              \\
\textbf{OVRL + SLING \cite{SLING}}            & 71.2                     & 54.1                     & 60.3          & 44.4                     & 43.0            & 29.1            & 58.2                     & 42.5          & 68.4            & 47.0            & 57.7             & 39.8             & 40.2            & 25.5            & 55.4                              & 37.4                              \\ \midrule
\textbf{\proposed (ours)} & \multicolumn{1}{l}{76.4} & \multicolumn{1}{l}{55.3} & \textbf{73.6} & \multicolumn{1}{l}{46.1} & \textbf{54.6}   & \textbf{30.2}   & \textbf{68.2}            & 43.9          & \textbf{75.3}   & \textbf{52.5}   & \textbf{70.9}    & \textbf{42.3}    & \textbf{51.0}   & \textbf{27.4}   & \multicolumn{1}{c}{\textbf{65.7}} & \multicolumn{1}{c}{\textbf{40.8}} \\ 
% \textbf{\proposed w/o Noise}            & 74.0                     & 65.5                    & 68.1          & 56.0                    & 52.4           & 41.1            & 64.8                     & 54.2          & 71.7           & 64.5            & 67.2             & 55.1             & 45.2            & 35.3            & 61.4  & 51.6                              \\ 
% \textbf{\proposed w/o Noise, Latent score}            &    61.6          &42.5                   & 58.3          & 38.1                   &   41.9       &   25.3          &      54.8     &  35.9  & 59.0                    & 39.5          & 57.0           & 35.5           & 37.9            & 21.5                 & 50.7  & 31.6                             \\ 

\bottomrule
\end{tabular}

%% file: tables/ablation.tex
% \begin{tabular}{@{}cccccccccccc@{}}
% \toprule
% \multirow{2}{*}{\textbf{\proposed }} & \multirow{2}{*}{\textbf{\begin{tabular}[c]{@{}c@{}}Graph \\ Exploration\end{tabular}}} & \multirow{2}{*}{\textbf{\begin{tabular}[c]{@{}c@{}}GT \\ Point Navi\end{tabular}}} & \multirow{2}{*}{\textbf{\begin{tabular}[c]{@{}c@{}}Oracle\\ STOP\end{tabular}}} & \multicolumn{2}{c}{\textbf{Easy}} & \multicolumn{2}{c}{\textbf{Medium}} & \multicolumn{2}{c}{\textbf{Hard}} & \multicolumn{2}{c}{\textbf{Overall}} \\ \cmidrule(l){5-12} 
%  &  &  &  & \textbf{SR} & \textbf{SPL} & \textbf{SR} & \textbf{SPL} & \textbf{SR} & \textbf{SPL} & \textbf{SR} & \textbf{SPL} \\ \midrule
% o & o & o & o & 85.5 & 54.5 & 87.6 & 57.1 & 72.2 & 39.8 & 81.8 & 50.5 \\
% o & x & o & o & 41.4 & 22.7 & 24.9 & 15.4 & 6.6 & 4.2 & 24.3 & 14.1 \\
% o & o & o & x & 71.0 & 45.4 & 66.8 & 43.9 & 45.2 & 27.7 & 61.0 & 39.0 \\
% x & o & o & x & 60.4 & 28.9 & 57.1 & 26.1 & 37.8 & 17.2 & 51.8 & 24.1 \\ \midrule
% o & o & x & x & 70.3 & 63.3 & 52.4 & 43.7 & 19.6 & 15.4 & 47.4 & 40.8\\ \bottomrule
% \end{tabular}

\begin{tabular}{@{}cccrrrrrrcc@{}}
\toprule
\multicolumn{1}{l}{} & \multicolumn{1}{l}{} & \multicolumn{1}{l}{} & \multicolumn{2}{c}{\textbf{Easy}}                                  & \multicolumn{2}{c}{\textbf{Medium}}                                & \multicolumn{2}{c}{\textbf{Hard}}                                  & \multicolumn{2}{c}{\textbf{Overall}} \\ \midrule
Noise       & \textbf{$F_{loc}$}      & \textbf{$F_{track}$}    & \multicolumn{1}{c}{\textbf{SR}} & \multicolumn{1}{c}{\textbf{SPL}} & \multicolumn{1}{c}{\textbf{SR}} & \multicolumn{1}{c}{\textbf{SPL}} & \multicolumn{1}{c}{\textbf{SR}} & \multicolumn{1}{c}{\textbf{SPL}} & \textbf{SR}      & \textbf{SPL}      \\
\midrule
\redx                    & \redx                    & -                    & 60.3 & 41.0 & 57.7 & 36.8 & 39.9 & 23.4 & 52.6 & 33.7     \\
\redx                    & \greencheck                    & -              & 72.8 & 61.1 & 67.7 & 48.9 & 48.8 & 31.1 & 63.1 & 47.0       \\
\greencheck   & \greencheck       & \redx                                  & 74.6 & 53.6 & 64.0 & 40.9 & 40.8 & 23.6 & 59.8 & 39.4       \\ \midrule
\greencheck                      & \greencheck     & \greencheck           & 75.3 & 53.9 & 72.2 & 44.2 & 52.8 & 28.9 & 66.9 & 42.3               \\ \bottomrule
\end{tabular}

%% file: tables/loc_exp.tex
\begin{tabular}{@{}ccc|cccc@{}}
\toprule
 & \multicolumn{2}{c|}{\textbf{(a) Camera Tracking + Mapping}} & \multicolumn{4}{c}{ \textbf{(b) Image-Based Localization}} \\ \midrule
 & \textbf{ATE RMSE (m)} & \textbf{Inference Time (s)} & \textbf{25cm Recall (\%)} & \textbf{50cm Recall (\%)} & \textbf{1m Recall (\%)} & \textbf{Inference Time (s)} \\ \midrule
\textbf{Raw Noise} & 0.877 & - & - & - & - & - \\
\textbf{MapNet \cite{MapNet}} & 0.332 & \textbf{0.104} & 8.6 & 15.4 & 21.2 & 0.027 \\
\textbf{Nice-SLAM \cite{nice-slam}} & 0.941 & 6.743 & 85.9 & 93.9 & 94.3 & 24.520 \\
\textbf{Nice-SLAM* \cite{nice-slam}} & \textbf{0.071} & 25.977 & \textbf{91.1} & 93.7 & 94.4 & 24.520 \\ \midrule
\textbf{RNR-Map (Ours)} & 0.108 & 0.271 & 76.6 & \textbf{99.2} & \textbf{99.5} & \textbf{0.018} \\ \bottomrule
\end{tabular}

% \begin{tabular}{@{}ccc@{}}
% \toprule
%            & ATE RMSE (m) & Time (Mapping + Tracking) (s) \\ \midrule
% Raw Noise  & 0.877        & -                                       \\
% MapNet     & 0.332        & \textbf{0.212}                                   \\
% NICE-SLAM  & 0.941        & 6.743                                   \\
% NICE-SLAM* & \textbf{0.071}        & 25.977                                  \\ \midrule
% \proposed (ours)     & 0.108        & 0.570                                   \\ \bottomrule
% \end{tabular}

%% file: tables/novel_loc_exp.tex
\begin{tabular}{@{}ccccc@{}}

\toprule
\textbf{Difficulty} & \textbf{OC 1}                           & \textbf{OC 2}      & \textbf{OC 3}                            & \textbf{OC 4}     \\ \midrule                       %\\
%                     & $5\% < \ x \leq 10\%$ & $10\% < \ x \leq 20\%$ & $20\% < \ x \leq 30\%$ & $30\%  < \ x \leq 40\%$ \\ 
\textbf{Avg Img. diff.}   & 7\%                               & 14.1\%                                                     & 23.8\%                             & 33.9\%                             \\  \midrule  
\textbf{Dist. Err.(m)}     & 0.083                             & 0.091                                                      & 0.111                              & 0.140                              \\
\textbf{25cm Recall (\%)}     & 76.7                              & 72.8                                                       & 71.5                               & 69.6                               \\
\textbf{50cm Recall (\%)}     & 99.8                              & 98.8                                                       & 98.6                               & 97.4                               \\
\textbf{1m Recal (\%)}       & 99.8                              & 99.0                                                         & 99.2                               & 98.5                               \\ \bottomrule
% \toprule
% NE       & CL (\%) & CLIP (\%) & SSIM (\%) \\ \midrule
% Random   & 27.6    & 86.0        & 69.4      \\
% NeRF-Map & 85.0     & 94.5      & 74.1      \\ \bottomrule

\end{tabular}

%% file: tables/NE_experiment.tex
\begin{tabular}{@{}ccccc@{}}
\toprule
\multicolumn{1}{c}{\backslashbox[40mm]{\textbf{Search Method}}{\textbf{Similarity (\%)}}} & \textbf{CLIP \cite{clip}} & \textbf{CL} & \textbf{SSIM} & \textbf{L1} \\ \midrule
\textbf{Random} & 88.23 & 30.91 & 78.66 & 88.44 \\
\textbf{Max CLIP \cite{clip}} & {\color[HTML]{9B9B9B} 100.0} & 61.36 & 84.58 & 90.48 \\
\textbf{Max CL} & 93.26 & {\color[HTML]{9B9B9B} 100.0} & \textbf{91.29} & 91.60 \\
\textbf{Max SSIM} & 93.75 & 66.98 & {\color[HTML]{9B9B9B} 100.0} & \textbf{93.40} \\
\textbf{Max Inv. L1} & 90.77 & 49.68 & 90.91 & {\color[HTML]{9B9B9B} 100.0} \\ \midrule
\textbf{RNR-Map (Ours) } & \textbf{94.52} & \textbf{82.9} & 90.93 & 91.41 \\ \bottomrule
\end{tabular}

%% file: tables/supp_ablation.tex
\begin{tabular}{@{}ccccccccccccc@{}}
\toprule
 &
  \multicolumn{4}{c}{\textbf{Gibson-Curved}} &
  \multicolumn{2}{c}{\textbf{Easy}} &
  \multicolumn{2}{c}{\textbf{Medium}} &
  \multicolumn{2}{c}{\textbf{Hard}} &
  \multicolumn{2}{c}{\textbf{Overall}} \\ \midrule
 &
  \textbf{Exloration Score} &
  \textbf{Latent Score} &
  \textbf{GT Point Navi} &
  \textbf{GT Stopper} &
  \textbf{SR} &
  \textbf{SPL} &
  \textbf{SR} &
  \textbf{SPL} &
  \textbf{SR} &
  \textbf{SPL} &
  \textbf{SR} &
  SPL \\ \midrule
\multirow{3}{*}{\begin{tabular}[c]{@{}c@{}}Ground-truth \\ Information\end{tabular}} 
& \greencheck & \greencheck & \greencheck & \greencheck & 98.0 & 85.0 & 95.0 & 70.8 & 86.4 & 53.9 & 93.1 & 69.9 \\
& \greencheck & \greencheck & \redx & \greencheck & 94.6 & 81.8 & 90.1 & 66.9 & 72.2 & 45.0 & 85.6 & 64.6 \\
& \greencheck & \greencheck & \greencheck & \redx & 72.4 & 61.2 & 65.2 & 47.1 & 55.2 & 33.2 & 64.3 & 47.1 \\ \midrule \midrule
\multirow{2}{*}{\begin{tabular}[c]{@{}c@{}}Exploration\\ Strategy\end{tabular}}      
& \greencheck & \redx & \redx & \redx & 59.0 & 39.5 & 57.0 & 35.5 & 37.9 & 21.5 & 51.3 & 32.2 \\
& \redx & \greencheck & \redx & \redx & 71.1 & \textbf{60.5} & 65.9 & {45.4} & 40.4 & 24.2 & 59.2 & {43.4} \\ \midrule
Full Model & \greencheck & \greencheck & \redx & \redx & \textbf{71.7} & 59.9 & \textbf{67.2} & \textbf{47.5} & \textbf{45.2} & \textbf{29.2} & \textbf{61.4} & \textbf{45.6} \\ \bottomrule
\end{tabular}

%% file: tables/supp_mp3d.tex
\begin{tabular}{@{}cccccccccc@{}}
\toprule
\multirow{2}{*}{\textbf{Training Dataset}} & \multirow{2}{*}{\textbf{Method}} & \multicolumn{2}{c}{\textbf{Easy}} & \multicolumn{2}{c}{\textbf{Medium}} & \multicolumn{2}{c}{\textbf{Hard}} & \multicolumn{2}{c}{\textbf{Overall}} \\ \cmidrule(l){3-10} 
 &  & \textbf{SR} & \textbf{SPL} & \textbf{SR} & \textbf{SPL} & \textbf{SR} & \textbf{SPL} & \textbf{SR} & \textbf{SPL} \\ \midrule
MP3D & \textbf{NRNS \cite{NRNS}} & 23.7 & 12.7 & 16.2 & 8.3 & 9.1 & 5.1 & 16.3 & 8.7 \\
MP3D & \textbf{SLING+NRNS \cite{SLING}} & 43.2 & 19.7 & 32.5 & 15.1 & 22.1 & 9.9 & 32.6 & 14.9 \\
Gibson & \textbf{OVRL \cite{OVRL}} & 16.9 & 8.3 & 25.8 & 13.5 & 14.3 & 7.0 & 10.6 & 4.6 \\
MP3D & \textbf{OVRL \cite{OVRL}} & 52.4 & 35.2 & 42.6 & 26.3 & 29.7 & 16.9 & 41.6 & 26.1 \\
Gibson & \textbf{SLING+OVRL \cite{SLING}} & 47.5 & 30.2 & 28.4 & 17.0 & 19.3 & 9.3 & 31.7 & 18.8 \\
MP3D & \textbf{SLING+OVRL \cite{SLING}} & \textbf{62.6} & 41.1 & 48.4 & 31.5 & 29.2 & 17.7 & 46.7 & 30.1 \\ \midrule
Gibson & \textbf{RNR-Map (ours) } & 58.42 & \textbf{49.01} & \textbf{50.35} & \textbf{35.57} & \textbf{34.38} & \textbf{22} & \textbf{47.7} & \textbf{35.5} \\ \bottomrule
\end{tabular}